\documentclass[lettersize,journal]{IEEEtran}
\usepackage{amsmath,amsfonts}
\usepackage{algorithmic}
\usepackage{algorithm}
\usepackage{array}
\usepackage[caption=false,font=normalsize,labelfont=sf,textfont=sf]{subfig}
\usepackage{textcomp}
\usepackage{stfloats}
\usepackage{url}
\usepackage{verbatim}
\usepackage{graphicx}
\usepackage{cite}
\usepackage{multirow}
\usepackage{enumitem}
\usepackage{color}
\begin{document}

\title{Bi-Directional Digital Twin Prototype Anchoring With Multi-Periodicity Learning for Few-Shot Fault Diagnosis}

\author{
  Pengcheng Xia, \emph{Member, IEEE},
	Zhichao Dong,
	Chengjin Qin,
	Qun Chao,
	Yixiang Huang, \emph{Member, IEEE},\\
	and Chengliang Liu
        % <-this % stops a space
\thanks{
  This work was supported in part by the National Natural Science Foundation of China under Grant 52405122 and Grant 52375109, in part by the National Key Research and Development Program of China under Grant 2023YFB3408502, and in part by the China Postdoctoral Science Foundation under Grant 2025M771378. \textit{(Corresponding author: Yixiang Huang.)}
		
	The authors	are with the State Key Laboratory of Mechanical System and Vibration, Shanghai Jiao Tong University, Shanghai 200240, China (e-mail: xpc19960921@sjtu.edu.cn; mezcdong@sjtu.edu.cn; qinchengjin@sjtu.edu.cn; chaoqun@sjtu.edu.cn; huang.yixiang@sjtu.edu.cn; chlliu@sjtu.edu.cn).

	This work has been accepted for publication in \textit{IEEE Internet of Things Journal}. Final publication is available at https://doi.org/10.1109/JIOT.2026.3748376.
	
	© 2026 IEEE.  Personal use of this material is permitted.  Permission from IEEE must be obtained for all other uses, in any current or future media, including reprinting/republishing this material for advertising or promotional purposes, creating new collective works, for resale or redistribution to servers or lists, or reuse of any copyrighted component of this work in other works.}% <-this % stops a space
}

% The paper headers
\markboth{IEEE INTERNET OF THINGS JOURNAL}%
{Shell \MakeLowercase{\textit{et al.}}: A Sample Article Using IEEEtran.cls for IEEE Journals}

% \IEEEpubid{0000--0000/00\$00.00~\copyright~2021 IEEE}
% Remember, if you use this you must call \IEEEpubidadjcol in the second
% column for its text to clear the IEEEpubid mark.

\maketitle

\begin{abstract}

Intelligent fault diagnosis (IFD) typically requires abundant labeled data, which are difficult to obtain from physical assets in practical industrial applications. Digital twins (DTs) provide a promising alternative by generating labeled simulation data, while most existing DT-assisted diagnosis methods rely on domain adaptation and still require considerable unlabeled target-domain data. To enable diagnostic knowledge transfer when only a few labeled target samples are available, this paper proposes a bi-directional DT prototype anchoring framework with multi-periodicity learning for few-shot fault diagnosis. The model is first meta-trained in the DT virtual space to acquire transferable diagnostic knowledge and is then adapted to the physical asset through test-time adaptation. To improve cross-domain adaptation under extremely limited target data, a bi-directional twin-domain prototype anchoring strategy is developed to mutually constrain the source- and target-domain class representations, while covariance-guided augmentation is introduced to improve the robustness of target prototype estimation. In addition, a multi-periodicity feature learning module is designed to capture the intrinsic periodic characteristics of motor current signals. A finite-element-based DT of an asynchronous motor is constructed for experimental validation under three rotational speeds and multiple few-shot settings. The experimental results show that the proposed method outperforms representative comparison methods, while the ablation studies verify the effectiveness of the components. These results demonstrate the potential of the proposed framework for DT-to-asset diagnostic knowledge transfer under severely limited target-domain data.
\end{abstract}


\section{Introduction}
\label{sec:introduction}
\IEEEPARstart{F}{ault} diagnosis plays a critical role in guaranteeing the reliability, safety, and availability of industrial systems, directly influencing production efficiency and operational sustainability~\cite{Motor_faults}. Consequently, developing advanced and reliable fault diagnosis methods to accurately identify faults from internet of thing (IoT) data has become an urgent and indispensable task in modern industrial systems~\cite{Motor_survey}.

In recent years, driven by the rapid advancement of artificial intelligence, intelligent fault diagnosis (IFD) has emerged as a powerful paradigm and achieved substantial breakthroughs~\cite{review2}. IFD methods are capable of automatically learning discriminative fault-related representations directly from raw monitoring signals from industrial internet of things (IIoT) systems~\cite{Data-driven, IIoT}. Owing to their strong representation learning capability, 1D convolutional neural networks (CNNs) have been extensively employed to extract hierarchical features from vibration, current, and other sensory data~\cite{CNN2}. Some studies transformed 1D signals into 2D time-frequency representations and employed the Vision Transformer (ViT) model to extract high-level features for diagnostic purposes~\cite{SDViT}.

Despite the impressive performance achieved by existing motor IFD approaches, they rely on the availability of abundant labeled data for model training. However, in practical industrial scenarios, collecting large-scale fault samples for a machine is often challenging~\cite{Small_sample}. Although data augmentation~\cite{DA1,DA2} can increase data volume, limited additional information is introduced, restricting model robustness. Training a diagnostic model using data collected from other operating conditions has emerged as a practical solution to alleviate the data scarcity issue~\cite{TL3}. In particular, domain adaptation (DA) techniques~\cite{TL5} have been widely exploited to align data distributions across different working conditions by explicitly minimizing statistical distribution differences using metrics such as maximum mean discrepancy (MMD)~\cite{MMD} and local MMD (LMMD)~\cite{TL4} or implicitly learn domain-invariant feature representations through domain adversarial learning~\cite{Adversarial}. In addition, meta-learning-based approaches have also been explored, aiming to train models with the ability to rapidly adapt to new diagnostic tasks using only a few labeled samples. For instance, prototypical networks (ProtoNets) have been introduced to realize few-shot diagnosis under previously unseen working conditions~\cite{motor}. {More recently, Xue et al.~\cite{MWAPN} proposed a multi-scale wavelet attention prototype network with adaptive region-based prototypes to enhance robustness under cross-domain distribution shifts. Wen et al.~\cite{RPN-CDM} developed a retrospective prototype network with a center difference measure to improve prototype representation and similarity evaluation for cross-machine diagnosis. Existing methods primarily investigate few-shot knowledge transfer across physical machines, components, or operating conditions.}

Nevertheless, several challenges still hinder the practical applications of IFD methods based on transfer learning or meta-learning frameworks. First, collecting a large amount of labeled training data from the source domain that comprehensively covers various fault categories is both time-consuming and resource-intensive. In addition, if the source-domain data exhibit substantial distribution discrepancies from the target domain, negative transfer may occur, leading to performance degradation~\cite{TL2}. With the rapid development of digital twin (DT) technology, virtual models that accurately replicate the physical behavior of machinery have provided a promising alternative for data acquisition with lower cost and potentially improved similarity~\cite{DT,DT-UAV}. As digital representations of physical machinery, discrepancies inevitably exist between DT-generated data and real-world measurements. Therefore, DA plays a crucial role in bridging the gap between virtual and physical domains to ensure reliable performance on actual assets~\cite{DT-review}. For example, Liu et al.~\cite{DT2} employed an unsupervised domain adaptation framework based on adversarial learning to transfer diagnostic knowledge from simulated bearing signals to real measurements. Zhang et al.~\cite{DT3} proposed a partial domain adaptation network to facilitate knowledge transfer from dynamics-based DT data of bearings. Xia et al.~\cite{DT4} constructed a motor DT model using the finite element method (FEM) and developed a semi-supervised framework to enhance cross-domain knowledge transfer. Beyond conventional TL strategies, Xia et al.~\cite{DT6} proposed a knowledge distillation framework from a motor DT to the physical asset with improved interpretability. Ma et al.~\cite{DT-MTL} integrated meta-learning and TL to overcome discrepancies between simulation and measurement data. Collectively, these studies have partially alleviated the dependency on source-domain data availability.

However, it should be noted that these DT-driven approaches still rely on the availability of a considerable amount of unlabeled target-domain samples for DA or knowledge transfer. In practical industrial environments, collecting sufficient target-domain data that comprehensively cover various fault information is often infeasible. A more realistic scenario is that only a very limited number of samples can be acquired and annotated before system shutdown. Under such circumstances, few-shot fault diagnosis represents a more practical yet significantly more challenging setting. Nevertheless, how to effectively leverage DT data to support few-shot learning in the physical domain has received limited attention. Therefore, developing a few-shot fault diagnosis framework that fully exploits DT knowledge remains an open and critical research problem.

Furthermore, most existing DTs generate vibration signals of rotating machinery. As for electromechanical machines such as asynchronous motors, current signals can be simulated with higher fidelity, as they are less susceptible to interference from external mechanical components and environmental disturbances. Furthermore, current signals are easier to acquire in industrial applications where vibration sensors cannot be mounted directly on the electromechanical machines~\cite{motor_current}. Nevertheless, current signals typically exhibit strong inherent periodicity, which may limit the effectiveness of feature extraction of CNNs due to their limited receptive fields, especially under few-shot conditions. Consequently, how to directly process the 1D current signals and cope with the strong periodicity remains to be explored.

To address the aforementioned challenges, this paper proposes a bi-directional DT prototype anchoring method with multi-periodicity learning for few-shot fault diagnosis. First, a framework involving meta-training in the virtual space and test-time adaptation in the physical space is proposed, enabling few-shot adaptation to the target machine. Second, a bi-directional prototype anchoring strategy is developed to overcome the data discrepancies between DT domains and {promote mutual feature alignment}, and a covariance-guided augmentation mechanism is also introduced for enhanced robustness. Furthermore, a multi-periodicity feature learning module is designed to capture temporal variations within and across multiple periodic cycles of current signals. We used an asynchronous motor as a case study and constructed its DT model through FEM and electromagnetic FEM simulations. Few-shot diagnosis experiments under three different rotational speeds verified the effectiveness of the proposed method and its superiority compared to other advanced approaches. The main contributions of this paper are summarized as follows.
\begin{enumerate}
	\item A DT-assisted few-shot fault diagnosis framework based on meta-learning and test-time adaptation is proposed.
	\item To mitigate data discrepancies between digital and physical domains, a bi-directional twin-domain prototype anchoring strategy is developed for few-shot adaptation.
	\item A multi-periodicity learning module is introduced for efficient representation learning within and across periods from current signals.
\end{enumerate}

The rest of the paper is organized as follows. Section~\ref{sec:DT} gives the detailed description of the DT model of the asynchronous motor. Section~\ref{sec:Method} presents the proposed method, and Section~\ref{sec:Experiment} describes the experiments and results. Finally, Section~\ref{sec:Conclusion} concludes the paper.

\section{Digital Twin Modeling}
\label{sec:DT}
{The proposed method is primarily designed for rotating electrical machinery represented by asynchronous motors. In this paper, a three-phase squirrel-cage asynchronous motor is selected as the application object to validate the proposed DT-assisted few-shot diagnosis framework. The electromagnetic DT model of this motor was established based on FEM for high-fidelity data generation. The considered diagnostic scenario assumes that sufficient labeled fault samples can be generated from the motor DT, whereas only a few labeled measurements are available from the physical motor.}
\subsection{Physical Asset}
\begin{figure}
	\centering
	\includegraphics[width=0.8\linewidth]{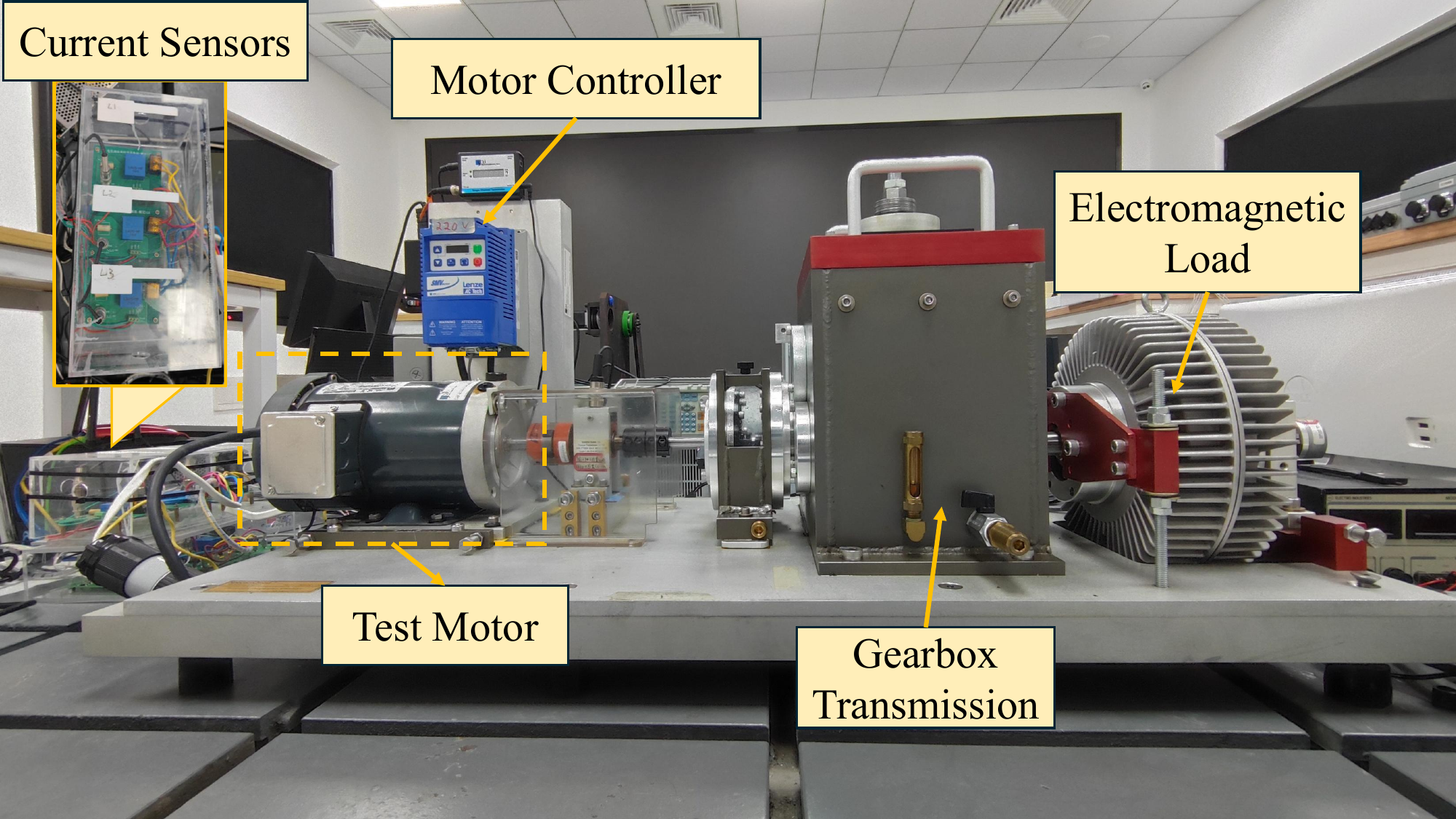}
	\caption{Experimental platform of the physical asset.}
	\label{fig:platform}
\end{figure}

Fig~\ref{fig:platform} presents the experimental platform of the physical motor. The platform is a Drivetrain Dynamics Simulator (DDS) platform, which primarily consists of the test motor and its controller, the gearbox transmission system, and the electromagnetic load attached to the gearbox. Three current sensors are mounted to collect the three-phase stator current signals of the test motor. The rotational and the load of the motor are controlled by the controller and electromagnetic load, respectively. The essential geometric dimensions and electrical parameters were carefully identified to ensure consistency of DT model, as shown in Table~\ref{tab:motor_param}. In experiments, we fixed the load applied to the motor at 2.4 N$\cdot$m. Experiments were conducted {at three rotational speeds} of 1200 r/min, 2400 r/min, and 2700 r/min.
\begin{table}
	\centering
	\caption{Primary geometric and electrical parameters of the physical motor.}\label{tab:motor_param}
	\setlength{\tabcolsep}{2pt}
	\begin{tabular}{cccc}
		\hline
		Parameter & Value & Parameter & Value\\
		\hline
		Stator outer diameter & 160.0 mm & Rated power & 2.2 kW\\
		Stator inner diameter & 81.3 mm & Rated speed & 3450 RPM\\
		Rotor outer diameter & 80.2 mm & Winding connection & Y-type\\
		Rotor inner diameter & 30.2 mm & Pole pairs & 1\\
		Stator slot number & 24 & Turns per stator winding slot & 92\\
		Rotor slot number & 34 & Parallel branch number & 4\\
		\hline
	\end{tabular}
\end{table}

\subsection{Virtual DT model}
\begin{figure}
	\centering
	\includegraphics[width=0.6\linewidth, trim=100 0 150 130, clip]{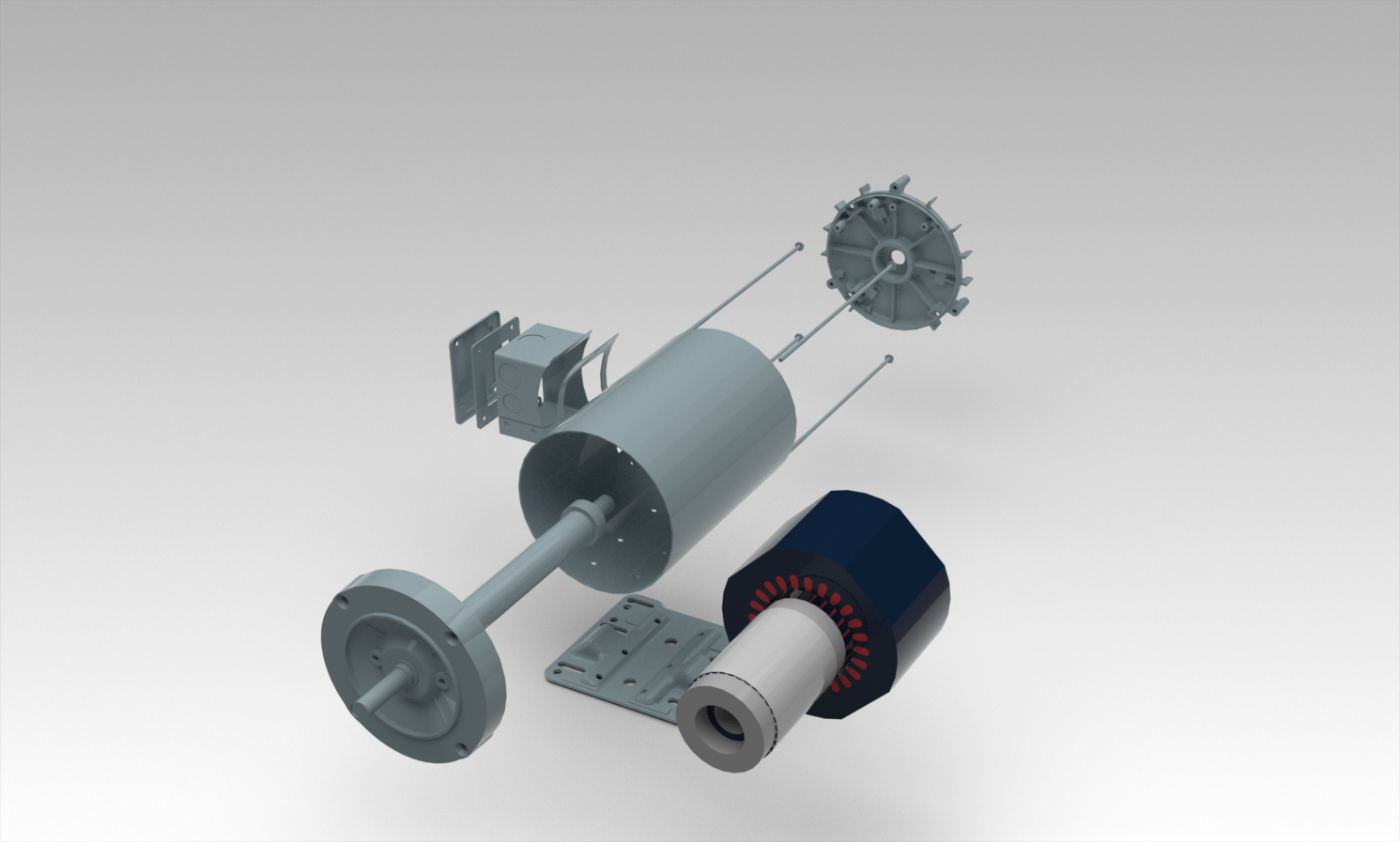}
	\caption{The disassembled view of the motor geometric model.}
	\label{fig:geometic}
\end{figure}
The DT was constructed by reconstructing the motor's geometric configuration according to its structural specifications, {where mechanical structures with negligible influence on the electromagnetic field were simplified to reduce computational complexity while retaining the dominant electromagnetic behavior relevant to stator-current generation.} The disassembled view of the geometric model is illustrated in Fig.~\ref{fig:geometic}.

Considering the relatively small axial variation of the electromagnetic field and the substantial computational burden associated with full three-dimensional simulations, a 2D FEM scheme was adopted~\cite{Simulation}. The cross-sectional plane perpendicular to the motor shaft was extracted to represent the electromagnetic domain. To emulate realistic operating conditions, the electromagnetic field model was coupled with an external circuit representing the power supply. A balanced three-phase voltage source was imposed, and stator windings were formulated within the circuit-field coupling framework. After motor startup, a constant mechanical load torque of 2.4 N$\cdot$m was applied to the shaft to simulate steady-state operation consistent with the physical experiment.

\begin{figure}
	\centering
	\includegraphics[width=0.7\linewidth]{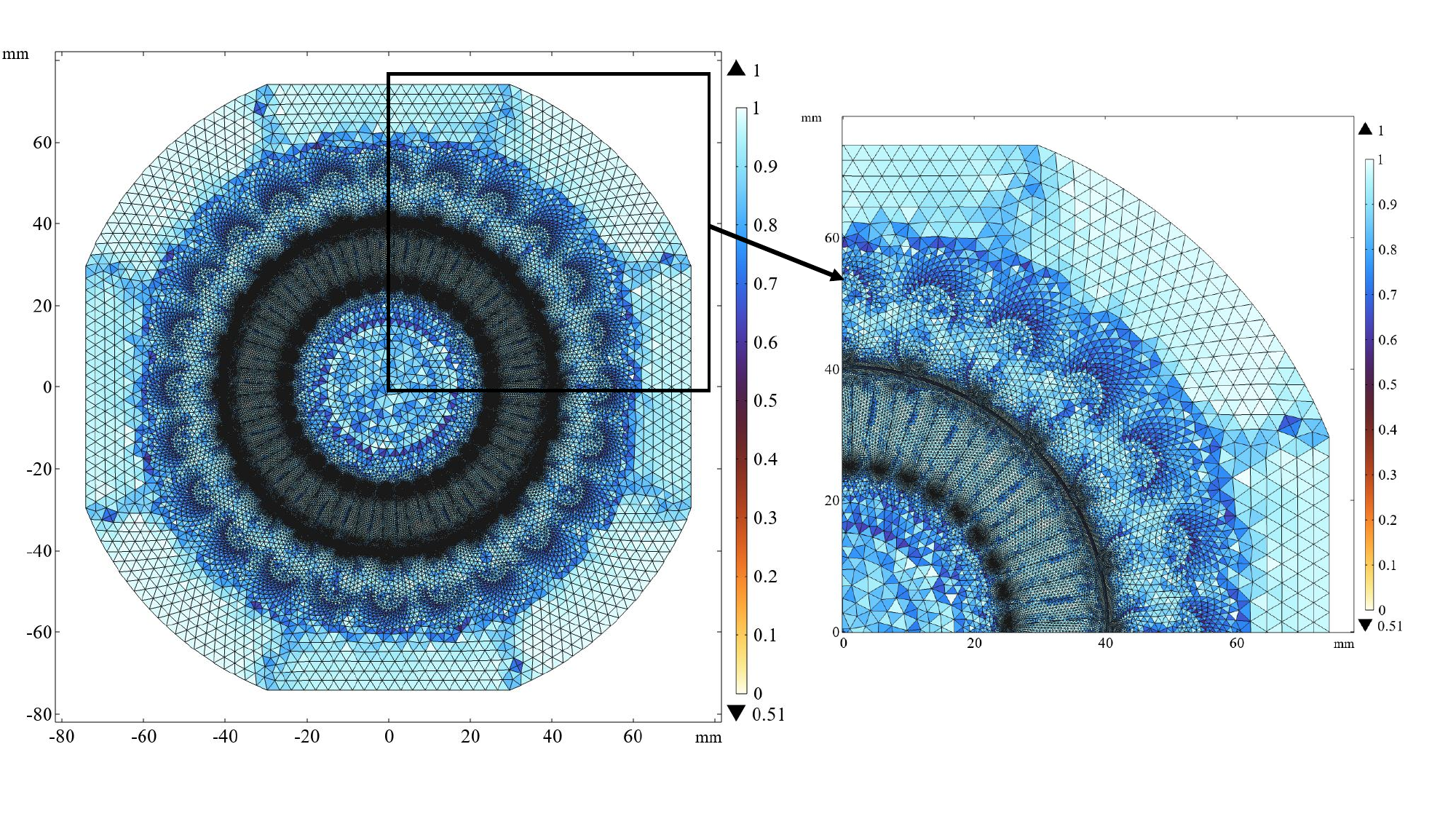}
	\caption{Mesh quality of the motor.}
	\label{fig:mesh}
\end{figure}

Mesh discretization was carefully designed to balance accuracy and efficiency. Regions with strong electromagnetic gradients, such as rotor bars and the air gap, were refined with dense triangular meshes with the maximum mesh sizes of 0.5 mm and 0.25 mm, respectively, whereas coarser meshes were employed {in the remaining, less magnetically sensitive regions}. The quality of the generated mesh model is depicted in Fig.~\ref{fig:mesh}. Simulations were conducted using a transient solver, and the stator current was sampled with a time step of 1/10240 second to ensure high-resolution temporal signals.

Three steady operating speeds corresponding to the physical experiments were simulated. Simulations were first performed under the healthy normal state (N) to establish baseline electromagnetic behavior. Fig.~\ref{fig:magnetic} illustrates the magnetic flux density and magnetic potential distributions of the normal motor at the rotational of 2700 r/min. Subsequently, representative motor faults were introduced by modifying structural or electrical parameters in the model, enabling the generation of simulated current signals under multiple faulty states. Three representative fault types commonly observed in asynchronous motors were incorporated into the DT model:
\begin{figure}
	\centering
	\subfloat[]{
		\includegraphics[width=0.45\linewidth]{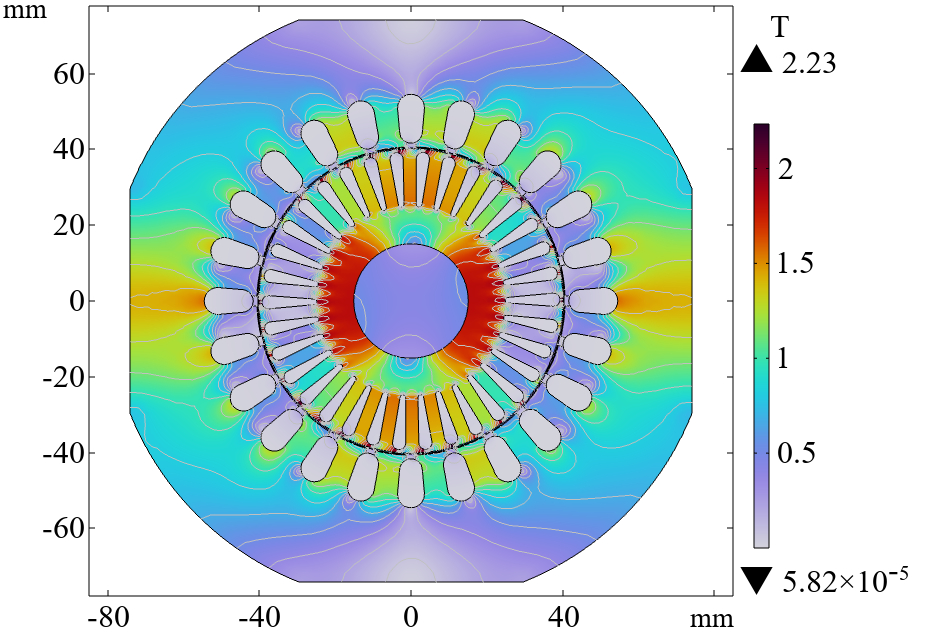}
	}
	\subfloat[]{
		\includegraphics[width=0.45\linewidth]{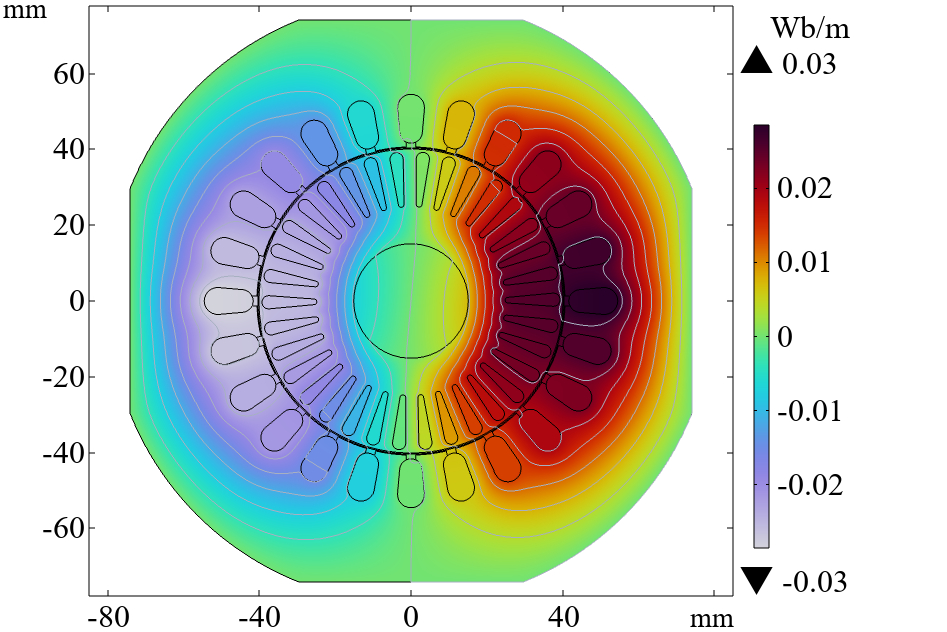}
	}
	\caption{(a) The magnetic flux density distribution and (b) the magnetic potential distribution of the normal motor.}
	\label{fig:magnetic}
\end{figure}

\begin{figure}
	\centering
	\subfloat[]{
		\includegraphics[width=0.40\linewidth]{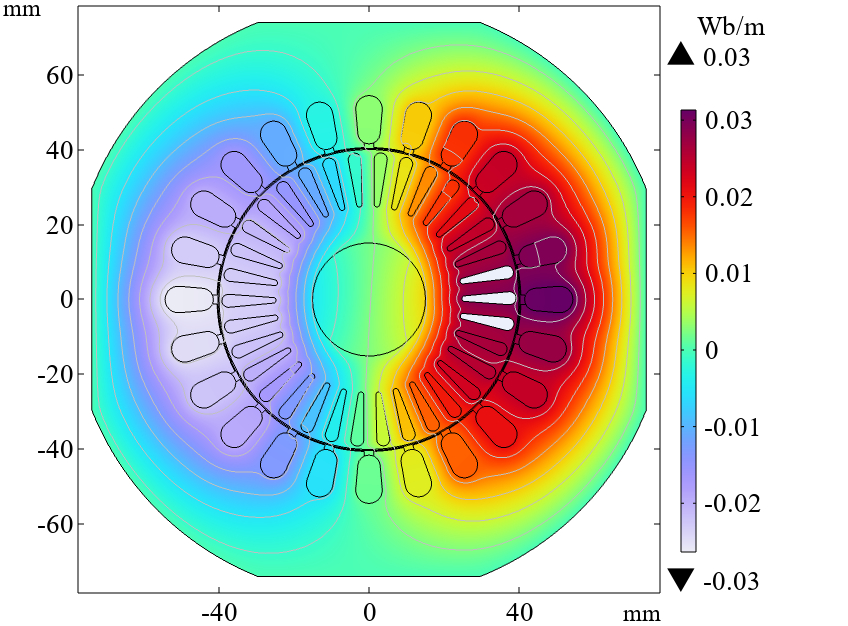}
	}
	\subfloat[]{
		\includegraphics[width=0.40\linewidth]{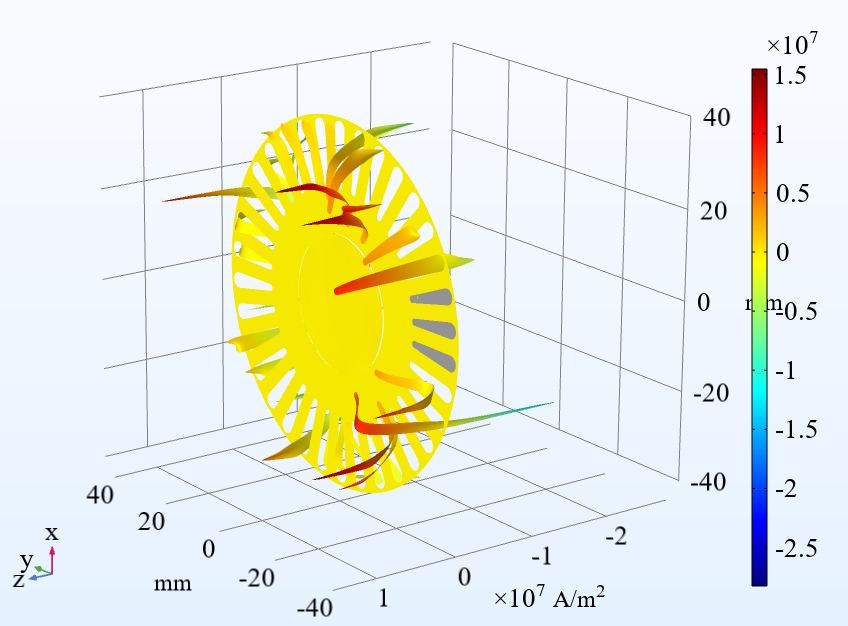}
	}
	\caption{(a) The magnetic potential distribution and (b) the current density distribution of the BRB motor.}
	\label{fig:BRB}
\end{figure}

\begin{figure}
	\centering
	\includegraphics[width=0.7\linewidth]{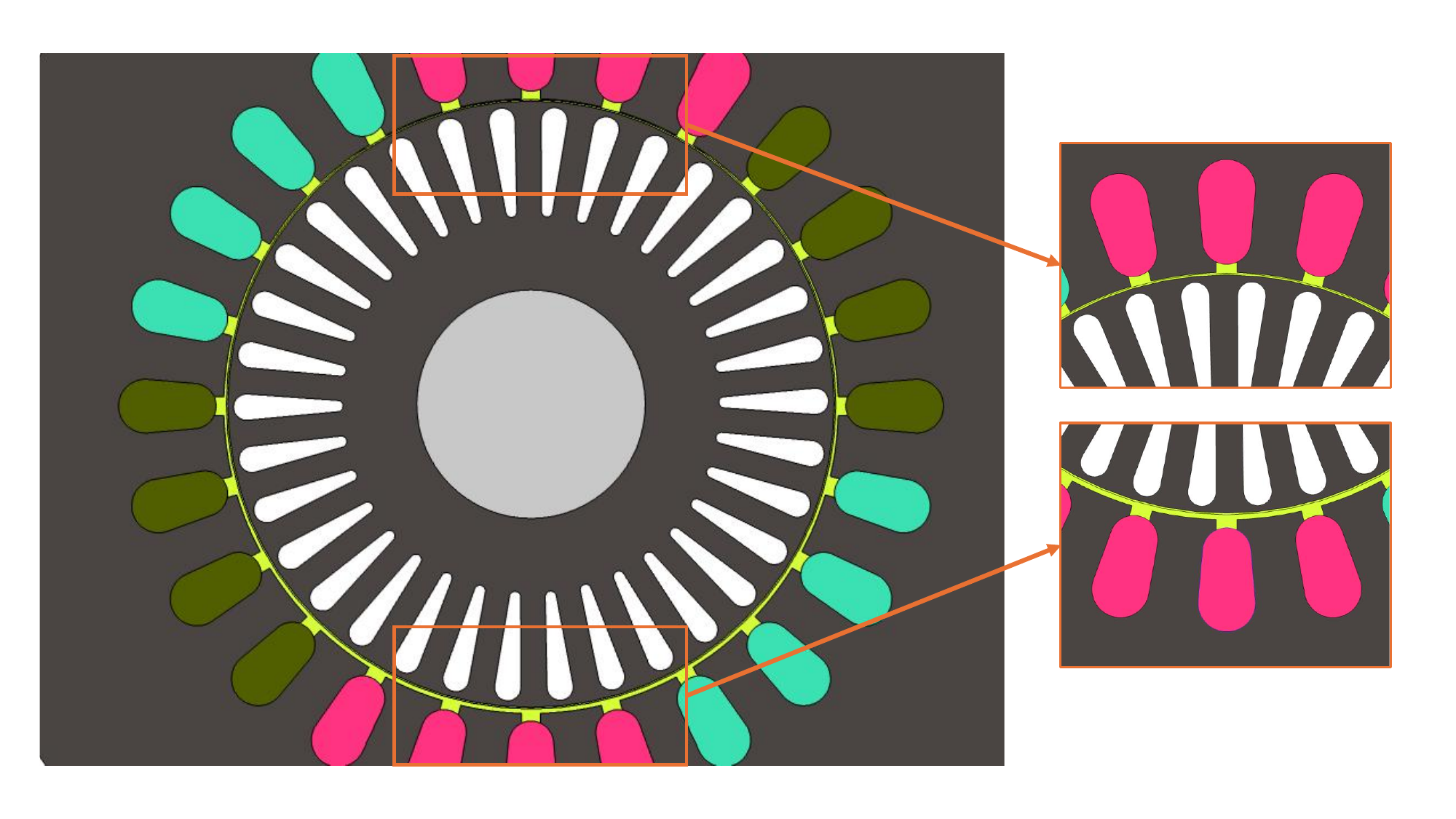}
	\caption{Geometry illustration of the MRF motor.}
	\label{fig:MRF}
\end{figure}

\begin{enumerate}[leftmargin=*]
	\item\textit{Broken Rotor Bar (BRB)}: Rotor bar breakage was emulated by altering the electrical conductivity of selected rotor bars. Instead of completely removing the bars from the geometry, their conductivity was drastically reduced to a near-insulating level (1 S/m), thereby suppressing current conduction while preserving structural continuity. Fig.~\ref{fig:BRB} depicts the magnetic potential distribution and the current density in rotor bars of the BRB motor. It can be observed that nearly no magnetic potential existed within the broken rotor bars, and the affected bars exhibited negligible current density, whereas adjacent bars carried redistributed currents due to electromagnetic coupling effects.
	\item\textit{Stator Winding Fault (SWF)}: An inter-turn short-circuit fault was modeled by modifying the electrical parameters of one phase winding. Specifically, the resistance of a selected winding in phase A was reduced by 5\%, representing partial insulation degradation.
	\item\textit{Misaligned Rotor Fault (MRF)}: Static rotor eccentricity was simulated through geometric perturbation. A radial displacement of 0.01 inches was imposed on the rotor, creating a non-uniform air-gap distribution as shown in Fig.~\ref{fig:MRF}.
\end{enumerate}

\begin{figure}
	\centering
	\subfloat[]{
		\includegraphics[width=0.4\linewidth]{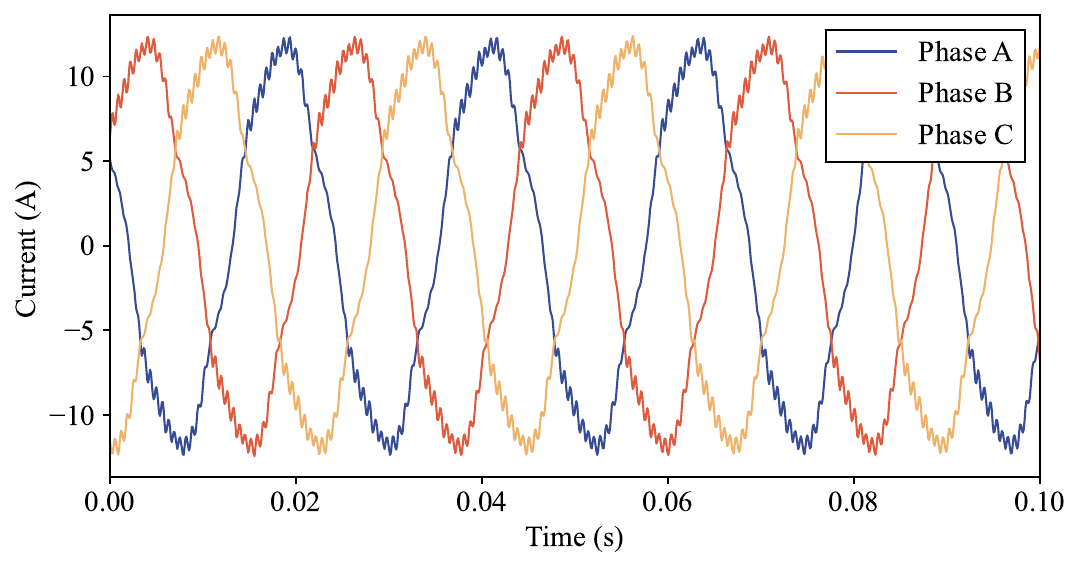}
	}
	\subfloat[]{
		\includegraphics[width=0.4\linewidth]{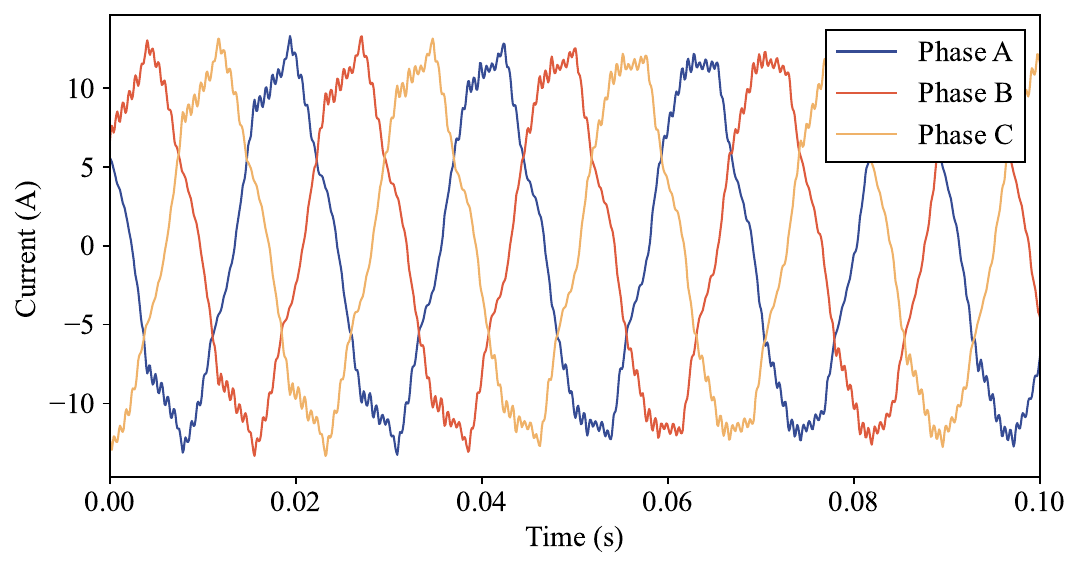}
	}
	\hfill
	\subfloat[]{
		\includegraphics[width=0.4\linewidth]{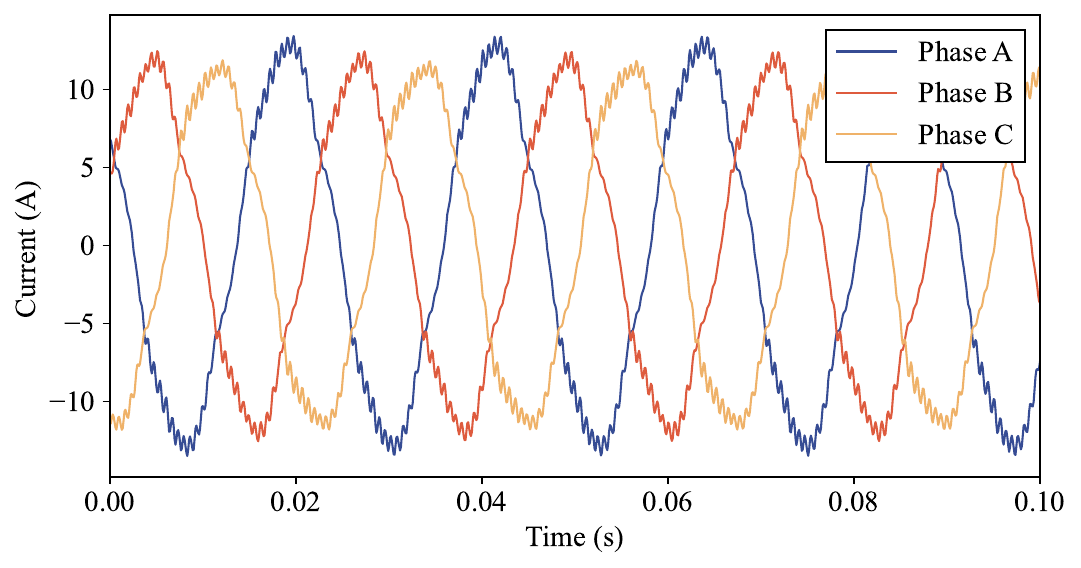}
	}
	\subfloat[]{
		\includegraphics[width=0.4\linewidth]{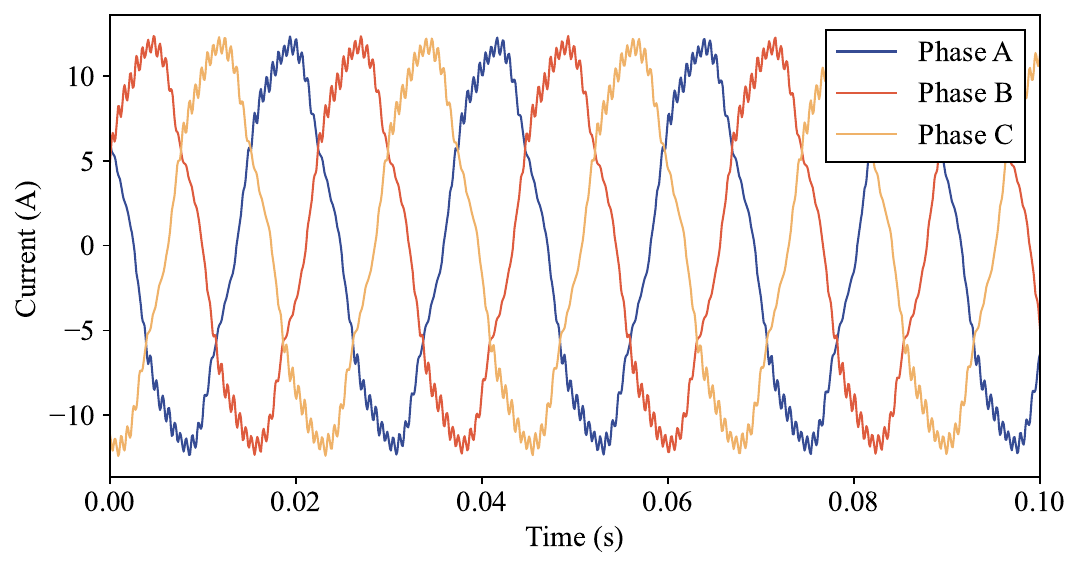}
	}
	\caption{Segments of simulated signals of motors under each health state. (a) N. (b) BRB. (c) SWF. (d) MRF.}
	\label{fig:signal}
\end{figure}

Following fault injection, three-phase stator currents were extracted from the coupled circuit-field model. The simulated current signals under each health state at the rotational of 2700 r/min are depicted as Fig.~\ref{fig:signal}. The healthy motor exhibits balanced sinusoidal currents with consistent amplitudes across phases. In contrast, the SWF state produces pronounced phase imbalance, while the BRB fault introduces mild waveform distortion. For the MRF case, waveform differences are relatively inconspicuous in the time domain.

{
\subsection{Twin-to-Asset Signal Fidelity Validation}
\label{subsec:fidelity_validation}

Before using the DT-generated currents to assist fault diagnosis of the real asset, their fidelity was quantitatively evaluated against the measurements collected from the physical motor. The comparison was conducted under the representative operating condition of 2700~r/min and 2.4~N$\cdot$m. Fault-frequency evidence and time-domain waveform consistency are considered to evaluate comprehensively.

\subsubsection{Fault-frequency evidence}

The fault-related frequency components were first examined to determine whether the injected faults produced physically meaningful signatures in the simulated stator currents. The theoretical frequencies were calculated independently according to the motor operating parameters and the corresponding fault mechanisms~\cite{fault_frequency}. As shown in Fig.~\ref{fig:fault_frequency_validation}(a)--(c), spectral enhancement can be observed around the theoretical components for all three simulated fault types.
\begin{figure*}
	\centering
	\subfloat[]{
		\includegraphics[width=0.3\linewidth]{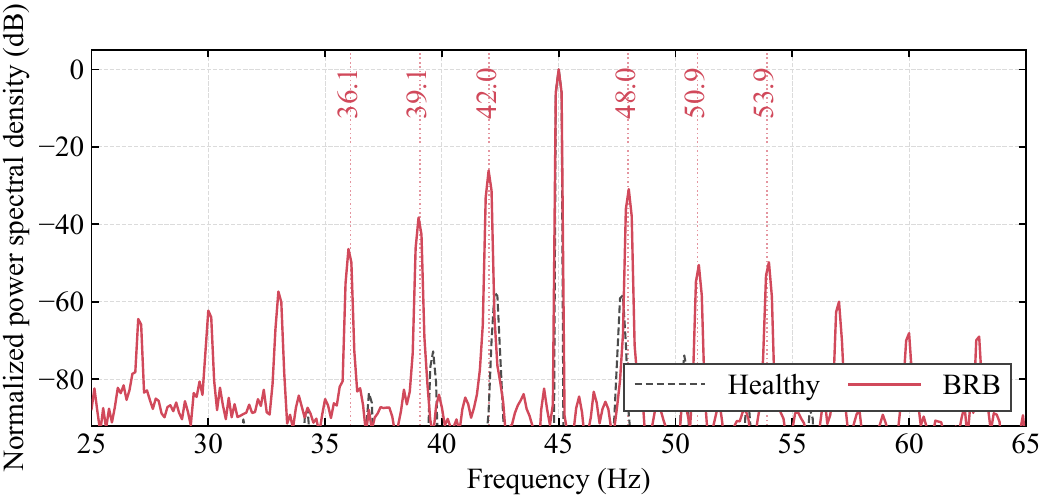}
	}
	% \hfill
	\subfloat[]{
		\includegraphics[width=0.3\linewidth]{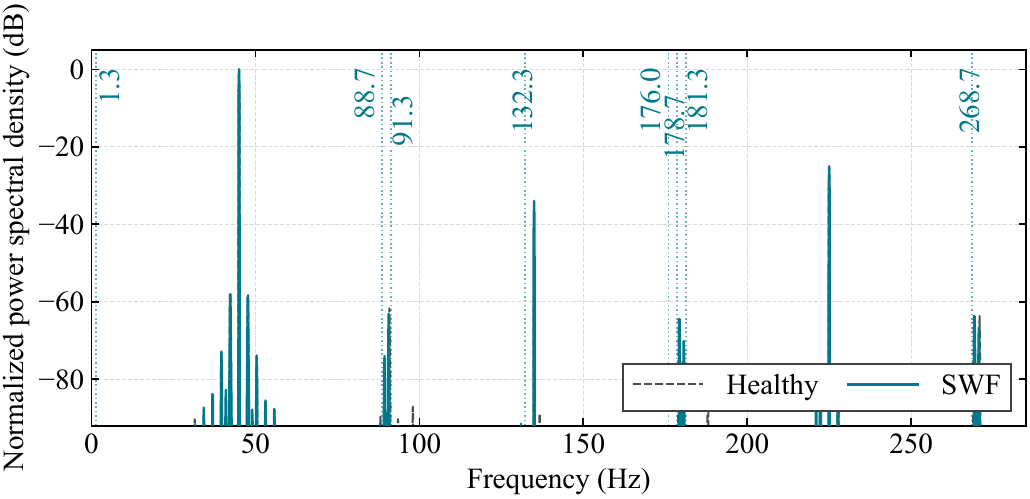}
	}
	% \hfill
	\subfloat[]{
		\includegraphics[width=0.3\linewidth]{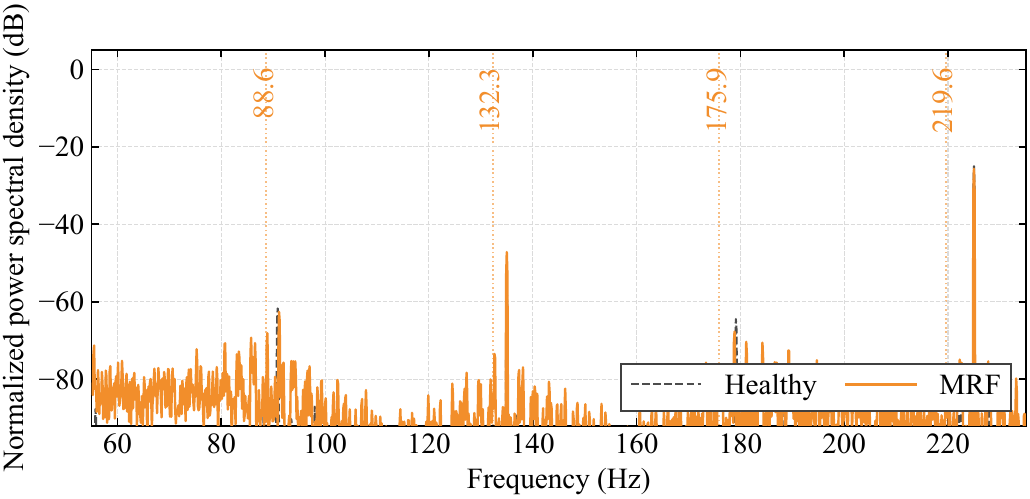}
	}
	\caption{{The normalized power spectral densities of the simulated healthy and faulty currents, where the vertical dotted lines denote the theoretically expected fault frequencies. (a) BRB. (b) SWF. (c) MRF.}}
	\label{fig:fault_frequency_validation}
\end{figure*}

To enable a direct comparison between the simulated and measured signals, the fault-to-healthy spectral gain was calculated at every theoretical component and then averaged within each fault category. The mean gains obtained from the simulated signals are 73.5~dB for BRB, 8.9~dB for SWF, and 53.8~dB for MRF, respectively. Therefore, positive spectral enhancement at the theoretically expected fault frequencies is exhibited. These results demonstrate that the DT reproduces the fault-dependent frequency components.

\subsubsection{Time-domain waveform consistency}

For the time-domain analysis, 40 simulated-measured sample pairs were evaluated for each health state. Per-sample normalization was performed, and a zero-mean normalized cross-correlation was first calculated for phase A, then the resulting lag was jointly applied to all three phases to preserve the inter-phase relationships. The median three-phase normalized cross-correlation coefficients between the simulated-measured sample pairs were calculated for each health state, and the results are 0.9937, 0.9929, 0.9730, and 0.9935 for N, BRB, SWF, and MRF, respectively. The corresponding normalized root mean square errors (NRMSE) are 0.0440, 0.0398, 0.0826, and 0.0465. Therefore, the DT reproduces the normalized waveform shapes well, although the SWF currents exhibit a relatively larger waveform discrepancy.

Overall, the analysis shows that the DT data effectively reproduces the current waveform shapes and the physically meaningful fault signatures, but not all signal details of the physical asset. The DT data are therefore suitable for providing source-domain diagnostic knowledge rather than directly replacing measured data. The remaining simulation-to-real discrepancy also quantitatively motivates the proposed test-time bi-directional prototype anchoring strategy.
}

\section{Methodology}
\label{sec:Method}
\subsection{Problem Formulation and Method Overview}
In this paper, a few-shot fault diagnosis problem is investigated, {in which only a few labeled samples from each health state of the target asset are available to obtain a reliable diagnosis model}. Let $\mathcal{D}_t$ denote the target domain consisting of real measured signals from the target machine, which provides only a few labeled samples for each class. Specifically, the support set is defined as
\begin{equation}
	\mathcal{S} = \{(\mathbf{x}_i, y_i)\}_{i=1}^{N\times K},
\end{equation}
where $N$ represents the number of classes and $K$ denotes the number of labeled samples per class, formulating an $N$-way $K$-shot problem. The support set of the target domain therefore contains $N\times K$ labeled samples. The objective of the model $f(\cdot)$ is to diagnose the fault categories of a query set $\mathcal{Q} = \{(\mathbf{x}_j, y_j)\}_{j=1}^{M}$ with only access to the support set.

In this paper, the DT of the physical asset is built and simulation data under each health state are generated, constructing a source domain with $K_s$ labeled samples:
\begin{equation}
	\mathcal{D}_s=\{(\mathbf{x}_k^s, y_k^s)\}_{k=1}^{K_s}.
\end{equation}
It should be noted that all the health states in the target domain need to be simulated, resulting in a shared label space of both domains, i.e., $\mathcal{Y}_s=\mathcal{Y}_t=\{1, 2, \ldots, N\}$. However, the data distributions of the two domains are different due to modeling inaccuracies and measurement noise, i.e., $P_s(X)\neq P_t(X)$. Consequently, the objective of this paper is to exploit the DT source domain data to assist model training.

\begin{figure}
	\centering
	\includegraphics[width=\linewidth]{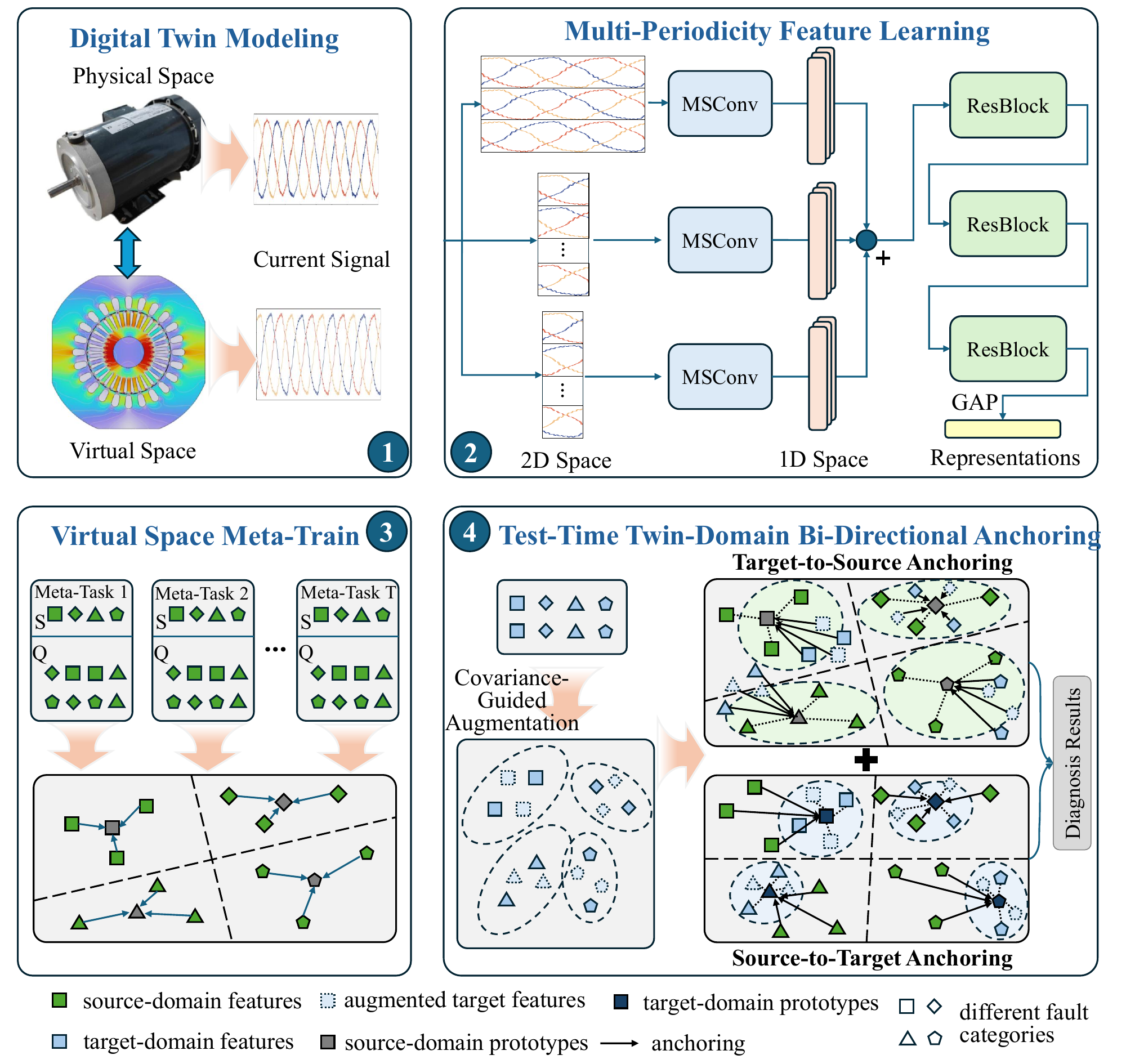}
	\caption{Framework illustration of the proposed method.}
	\label{fig:framework}
\end{figure}

{
To address the above problem, a DT-assisted few-shot diagnosis framework consisting of four sequential stages is proposed, as illustrated in Fig.~\ref{fig:framework}. First, a physics-based DT of the target asset is constructed to generate labeled source-domain fault data, while the measured signals from the physical asset form the target domain. Second, the proposed multi-periodicity feature learning network is employed to extract fault-discriminative representations from both domains. Third, episodic meta-training is performed using the DT-generated data to obtain a transferable model initialization. Finally, during deployment, the model is adapted to the physical asset using only a few labeled target support samples through covariance-guided augmentation and bi-directional prototype anchoring, while the target query samples are held out for final diagnosis. In the bi-directional prototype anchoring stage, target-to-source adaptation constrains the target features using source-domain prototypes, whereas source-to-target adaptation uses target-domain prototypes to guide the source representations toward the physical-domain distribution. After adaptation, the target query samples are classified according to the updated target-domain prototypes. The detailed implementations of these modules are introduced in the following subsections.
}

\subsection{Multi-periodicity Learning}
The stator current signals of electromechanical systems typically exhibit strong periodic characteristics determined by the electrical frequency and its harmonics. Effectively capturing such periodic structures is essential for extracting discriminative fault-related features. Therefore, inspired by the temporal processing paradigm in the TimesNet~\cite{timesnet}, a multi-periodicity feature learning module is designed to explicitly model temporal variations within and across different periods.

\begin{figure}
	\centering
	\includegraphics[width=0.9\linewidth]{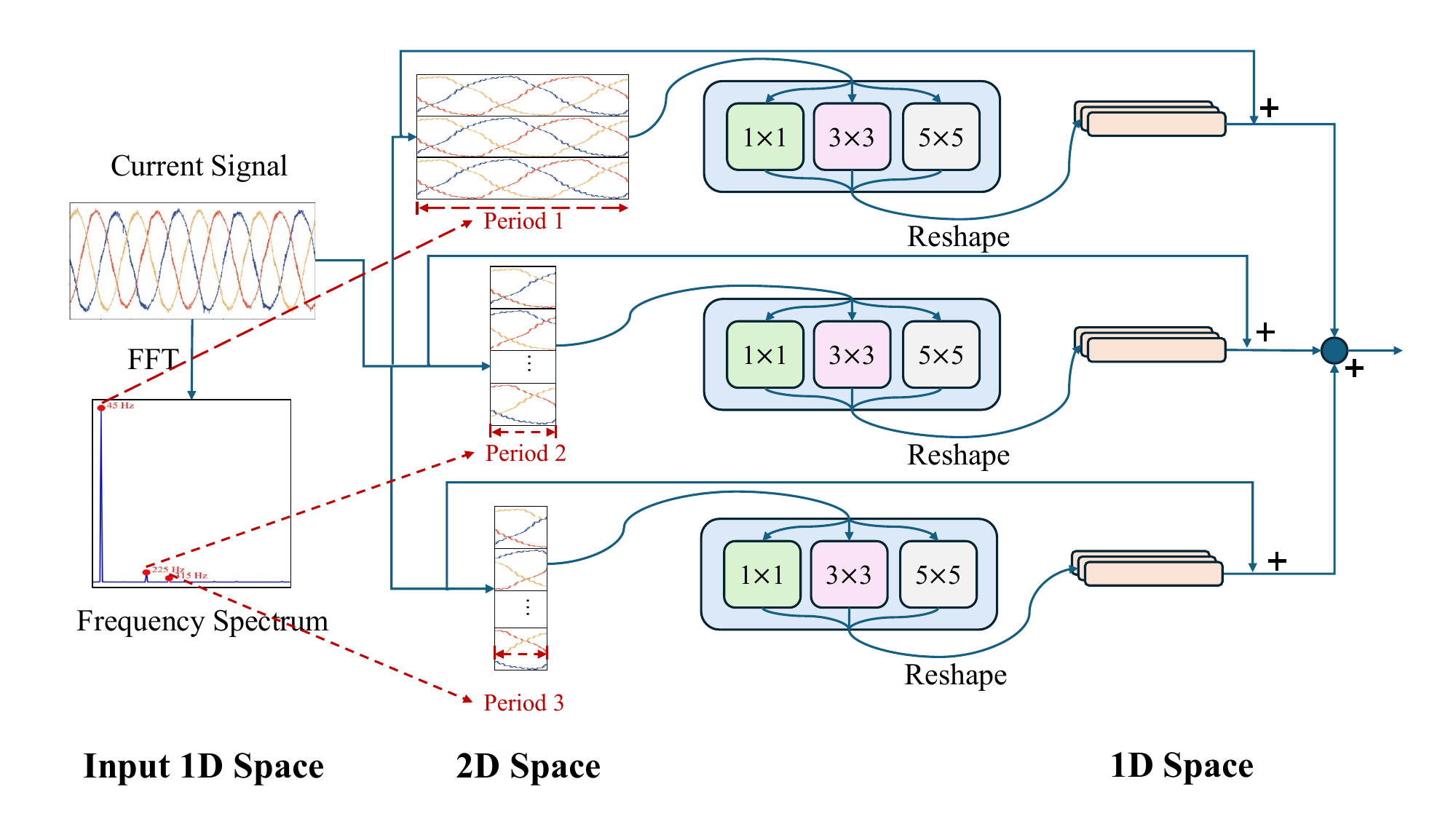}
	\caption{Network structure of the multi-periodicity learning module.}
	\label{fig:network}
\end{figure}

The structure of the multi-periodicity learning module is illustrated in Fig.~\ref{fig:network}. Specifically, the input consists of three-phase stator current signals $\mathbf{x}\in\mathbb{R}^{L\times 3}$, where $L$ denotes the signal length. First, the time-series signals are transformed into multiple 2D spaces according to their dominant periods. Instead of processing each phase independently, the frequency spectra of the three phases are first computed and then averaged to obtain a unified spectral representation as follows.
\begin{equation}
	A=\frac{1}{3}\sum_{c=1}^3\left\|\text{FFT}(\mathbf{x_c})\right\|,
\end{equation}
where $\mathbf{x_c}$ represents the $c$-th channel of the current signals, and $\text{FFT}$ denotes the single-sided frequency spectrum obtained by Fast Fourier Transform (FFT). Based on this averaged spectrum, the $k$ dominant frequency components are identified by selecting the frequencies with top-$k$ amplitudes, from which a set of candidate periods is determined:
\begin{equation}
	p_i=\lceil \frac{L}{f_i}\rceil,
\end{equation}
where $f_i$ is the $i$th dominant frequency.
\begin{figure}
	\centering
	\includegraphics[width=0.7\linewidth]{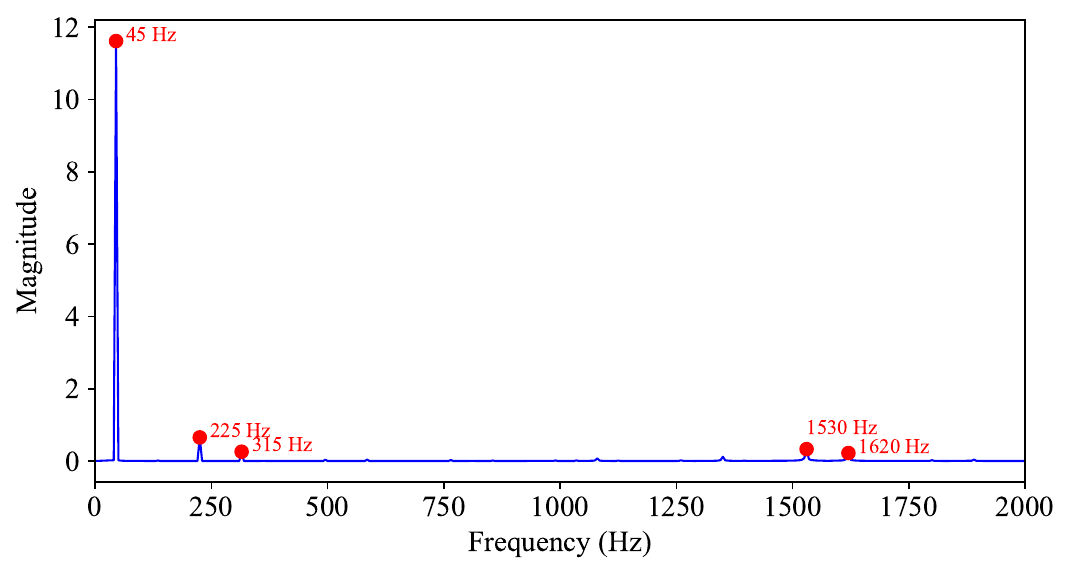}
	\caption{The frequency spectrum of the normal motor current signals and the top-5 dominant frequencies.}
	\label{fig:frequency}
\end{figure}

Taking the simulated current signals of the normal motor at rotational of 2700 r/min as an example, Fig.~\ref{fig:frequency} depicts the obtained spectrum and the top-5 dominant frequencies. It can be observed that the main dominant frequency is 45 Hz, corresponding to the rotational. The second and third dominant frequencies are 225 Hz and 315 Hz, {which are five and seven times the rotational frequency}, respectively. If we take 0.2 s as the time duration of samples, i.e., $L=2048$, the first three dominant periods are 46, 10, and 7, respectively. 

For each selected period, the current signals are reshaped into a 2D representation according to the corresponding periodic structure. This transformation converts long temporal dependencies into local spatial correlations, enabling convolutional operations to effectively capture both intra-period and inter-period patterns. Subsequently, a multiscale CNN (MSCNN) layer is designed to process the 2D representations from $k$ 2D spaces with multiple receptive fields. Three scales of $1\times1$, $3\times3$, and $5\times5$ are employed to extract features simultaneously and then aggregate them with average operation.

Subsequently, the obtained features of each period by the MSCNN layer are reshaped back into 1D spaces, respectively, leading to representations $\tilde{\mathbf{x}}_k\in\mathbb{R}^{L\times d}$, where $d$ denotes the channels of MSCNN. Residual connection is also designed to the original 1D signals, formulated as $\hat{\mathbf{x}}_k = \tilde{\mathbf{x}}_k + \mathbf{x}$. Afterward, features learned under multiple candidate periods are aggregated by the average operation, i.e., $\hat{\mathbf{x}} = \frac{1}{k}\sum_{i=1}^{k}\hat{\mathbf{x}}_i$. Through this process, the proposed module is able to capture diverse periodic patterns embedded in the current signals, thereby improving the robustness of feature extraction under limited data conditions. Subsequently, several residual blocks from ResNet~\cite{resnet} are stacked to further learn high-level representations from $\hat{\mathbf{x}}$, and a global average pooling (GAP) layer is applied to produce the final feature representation.

\subsection{Meta-training in Virtual Space}
After constructing the representation extraction network, meta-training is performed in the virtual space using the simulated data generated by the DT. The goal of this stage is to learn a transferable model that can rapidly adapt to the physical target domain with limited labeled samples.

Following the standard few-shot learning paradigm, the meta-training process is conducted in an episodic manner. Specifically, multiple meta-tasks are firstly designed, where each training episode mimics an $N$-way $K$-shot task consistent with the data settings in the target domain, while the query set includes additional $M$ samples used for evaluation within the episode. In each episode, the class prototype for each fault category {is computed by averaging the feature vectors of all the support samples following the ProtoNet~\cite{prototypical}:}
\begin{equation}\label{eq:proto}
	\mathbf{p}_n = \frac{1}{K}\sum\limits_{(\mathbf{x}_k, y_k)\in\mathcal{D}_t^s\atop y_k=n}f(\mathbf{x}_k),
\end{equation}
where $\mathcal{D}_t^s$ denotes the support set in the $t$th meta-task, $f(\cdot): \mathbb{R}^{L\times C}\rightarrow \mathbb{R}^w$ represents the representation extraction network, and $\mathbf{p}_n$ is the prototype of class $n$. For a query sample $\mathbf{x}_q$, its feature representation $f(\mathbf{x}_q)$ is compared with each class prototype using a distance metric $d(\cdot, \cdot)$, which is chosen as Euclidean distance in this paper. The probability that the query sample belongs to class $n$ is calculated as
\begin{equation}
	p(y_q=n|\mathbf{x}_q)=\frac{\exp \left[-d(f(\mathbf{x}_q), \mathbf{p}_n)\right]}{\sum_{n^\prime}\exp \left[-d(f(\mathbf{x}_q), \mathbf{p}_{n^\prime})\right]}.
\end{equation}
The model parameters are optimized by minimizing the negative log-probability of the real class over the query samples in each episode:
\begin{equation}\label{eq:protoloss}
	\begin{aligned}
	\mathcal{L}_s 
	&= \frac{1}{M}\sum_{n=1}^{N}\sum_{(\mathbf{x}_q,y_q)\in \mathcal{Q}_{t,n}^s}\Big[d(f(\mathbf{x}_q), \mathbf{p}_n)\\
	&\quad +\log \sum_{n^\prime}\exp \big(-d(f(\mathbf{x}_q), \mathbf{p}_{n^\prime})\big)\Big],
	\end{aligned}
\end{equation}
where $\mathcal{Q}_{t,n}^s$ denotes the subset with category $n$ of the query set in the $t$th meta-task. Through episodic meta-training on abundant DT-generated data, the model can form a strong initialization for subsequent test-time adaptation in the physical space with limited real-world samples.

\subsection{Test-time Twin-domain Model Adaptation}
Although meta-training in the virtual space enables the model to learn transferable representations, discrepancies inevitably exist between the simulated data and the measurements collected from the physical machine. In the few-shot diagnosis stage, a small support set $\mathcal{S}$ from the target machine is available, and traditional few-shot methods such as ProtoNet utilize the support set for query sample recognition with the well-trained model. However, the discrepancies may cause a shift in feature distributions, which degrades the reliability of the new prototypes when applied directly to the target domain. To address this issue, a test-time adaptation strategy is adopted using the small support set before model inference.

As the number of labeled samples in the target domain is extremely limited, an augmentation process is first conducted to alleviate the instability caused by few-shot observations. Instead of applying random perturbations, a covariance-guided augmentation strategy inspired by stochastic feature augmentation~\cite{augmentation} is adopted to generate more informative feature variations while preserving the intrinsic structure of each fault category. Specifically, given the extracted feature vectors $\mathbf{z} \in \mathbb{R}^{w}$ from the support samples, the features belonging to the same class are first grouped together. For each fault category $n$, the statistical distribution of the features is estimated by computing the covariance matrix based on the available samples. Let $\mathbf{z}_i^{n}$ denotes the feature vector of the $i$th sample in class $n$, and the covariance matrix $\mathbf{\Sigma}_n$ is obtained as
\begin{equation}\label{eq:cov}
	\mathbf{\Sigma}_n = \text{Cov}\left({\mathbf{z}_i^{n}}\right),
\end{equation}
which captures the variation patterns of the features within that class. It should be noted that the covariance matrix is assigned as an identity matrix in 1-shot scenario.

Based on the estimated covariance matrix, Gaussian perturbations are generated to create additional feature samples. For each original feature vector $\mathbf{z}$ with category $n$, $K_A$ augmented samples are generated by
\begin{equation}\label{eq:augmentation}
	\tilde{\mathbf{z}} = \mathbf{z}+\mathbf{\epsilon}, \mathbf{\epsilon}\sim\mathcal{N}(0,\mathbf{\Sigma}_n),
\end{equation}
where $\tilde{\mathbf{z}}$ denotes the augmented feature and $\mathbf{\epsilon}$ represents the class-dependent noise. In this way, the generated samples follow the statistical structure of the original features rather than arbitrary random noise, and additional feature samples with plausible intra-class variations can be obtained.

As discrepancies between data from two domains are inevitable, the prototypes of two domains are bound to be mismatched. Directly relying on either of the two prototypes may lead to suboptimal performance. On one hand, source-domain prototypes contain rich prior knowledge but may suffer from domain shift. On the other hand, target-domain prototypes better reflect the characteristics of the physical machine but are estimated from very limited samples and may be unstable. Consequently, similar to the feature alignment mechanism in DA~\cite{CDSSL}, a bi-directional prototype anchoring mechanism is proposed to align the representations of the virtual and physical domains during the adaptation.

Specifically, the source-domain prototypes firstly serve as knowledge anchors to regularize the target domain samples along with the augmented feature. This process can train the model to exploit the knowledge from DT model and prevent excessive deviation of target domain representations due to limited samples. First, class prototypes of the source domain are computed with source-domain samples as Equation (\ref{eq:proto}). For each fault category $n$, the source prototype is denoted as $\mathbf{p}_n^s$, which preserves the diagnostic knowledge learned from the DT during meta-training. Subsequently, the samples from the target domain are projected into the feature space and combined with the augmented features. The similarity between the target domain feature and each prototype $\mathbf{p}_n^s$ is measured using the Euclidean distance, and the target-to-source anchoring loss can be defined with the negative log-probability of the similarity with the corresponding prototype:
\begin{equation}
	\begin{aligned}
		\mathcal{L}_{anc1} 
		&= \frac{1}{N\times K}\sum_{n=1}^{N}\sum_{(\mathbf{x}_i,y_i)\in \mathcal{S}_n}\Big[d(f(\mathbf{x}_i), \mathbf{p}_n^s)\\
		&\qquad\qquad+\log \sum_{n^\prime}\exp \Big(-d\big(f(\mathbf{x}_i), \mathbf{p}_{n^\prime}^s\big)\Big)\Big]\\
		&\quad+ \frac{1}{N\times K\times K_A}\sum_{n=1}^{N}\sum_{\tilde{\mathbf{z}}_i\in \widetilde{\mathcal{Z}}_n}\Big[d(\tilde{\mathbf{z}}_i, \mathbf{p}_n^s)\\
		&\qquad\qquad+\log \sum_{n^\prime}\exp \Big(-d\big(\tilde{\mathbf{z}}_i, \mathbf{p}_{n^\prime}^s\big)\Big)\Big],
	\end{aligned}
\end{equation}
where $\widetilde{\mathcal{Z}}_n$ represents the augmented feature subset with category $n$ and {contains $K_A$ times as many samples as the support set}.

To further encourage confident predictions and more compact feature clustering, an entropy regularization term is further introduced. This term minimizes the entropy of the predicted class distribution as follows:
\begin{equation}\label{eq:entropy}
	\begin{aligned}
		\mathcal{L}_{ent1} 
		&= \mathbb{E}\Big[H\big[\frac{\exp \left[-d(f(\mathbf{x}), \mathbf{p}_n^s)\right]}{\sum_{n^\prime}\exp \left[-d(f(\mathbf{x}), \mathbf{p}_{n^\prime}^s)\right]}\big] \\
		&\quad + H\big[\frac{\exp \left[-d(\tilde{\mathbf{z}}_i, \mathbf{p}_n^s)\right]}{\sum_{n^\prime}\exp \left[-d(\tilde{\mathbf{z}}_i, \mathbf{p}_{n^\prime}^s)\right]}\big]\Big],
	\end{aligned}
\end{equation}
where $H(\cdot)$ denotes the entropy of class distribution.

The overall target-to-source adaptation loss is formulated as
\begin{equation}\label{eq:tsa}
	\mathcal{L}_{ts} = \mathcal{L}_{anc1} + \mathcal{L}_{ent1}.
\end{equation}

Conversely, the target-domain prototypes provide adaptation anchors that guide the representations from simulation data toward the distribution of real measurements. Similarly, the prototypes of all the target domain samples and the augmented features are firstly calculated, formulated as $\mathbf{p}_n^t$. Subsequently, the source-to-target anchoring loss is computed as follows:
\begin{equation}
	\begin{aligned}
	\mathcal{L}_{anc2} 
	&= \frac{1}{K_s}\sum_{n=1}^{N}\sum_{(\mathbf{x}_i,y_i)\in \mathcal{D}_n^s}\Big[d(f(\mathbf{x}_i), \mathbf{p}_n^t)\\
	&\quad +\log \sum_{n^\prime}\exp \left(-d(f(\mathbf{x}_i), \mathbf{p}_{n^\prime}^t)\right)\Big],
	\end{aligned}
\end{equation}
where $\mathcal{D}_n^s$ is the subset of source domain with sample category $n$. Subsequently, an entropy regularization term $\mathcal{L}_{ent2}$ is similarly obtained as Equation (\ref{eq:entropy}). The overall source-to-target adaptation loss can be presented as
\begin{equation}\label{eq:sta}
	\mathcal{L}_{st} = \mathcal{L}_{anc2} + \mathcal{L}_{ent2}.
\end{equation}

{The two adaptation directions are complementary. The source-domain prototypes are estimated from sufficient DT-generated samples and thus provide relatively stable class-wise diagnostic knowledge, whereas they inevitably contain a simulation-to-measurement domain bias. Therefore, the target-to-source adaptation constrains the few target samples using the stable source prototypes, preventing the feature extractor from being excessively adapted to unreliable observations in extremely few-shot scenarios. However, using only this direction may retain the source-domain bias and restrict the representations from sufficiently adapting to the physical asset. Conversely, the target-domain prototypes characterize the actual physical machine and therefore provide valuable anchors for correcting the domain shift. The source-to-target adaptation consequently guides the source representations toward the target class distributions. Nevertheless, these target prototypes are estimated from only a few measured samples and may exhibit relatively high estimation uncertainty. Relying solely on them may therefore result in unstable adaptation. By jointly minimizing $\mathcal{L}_{ts}$ and $\mathcal{L}_{st}$, the proposed bi-directional anchoring imposes mutual class-wise constraints between the two domains. Therefore, the overall test-time adaptation loss can be expressed as
\begin{equation}\label{eq:overall_loss}
	\mathcal{L}_t = \mathcal{L}_{ts} + \mathcal{L}_{st}.
\end{equation}
The adaptation therefore seeks a balanced feature space in which corresponding fault classes from the two domains become mutually consistent, rather than forcing one domain to completely follow the other. This mutual constraint also prevents either domain from dominating the adaptation process.}
After prototype anchoring, the query samples from the target domain are projected into the adapted feature space, and their class probabilities are computed based on the distances to the new prototypes of the target domain.

{
\subsection{Implementation Procedure}
The complete implementation of the proposed method consists of offline meta-training and online task-specific adaptation. First, the representation extractor is episodically meta-trained using the DT-generated source-domain data according to Eqs.~(\ref{eq:proto})--~(\ref{eq:protoloss}). For a new target diagnostic task, the meta-trained model first extracts the features of the few labeled target support samples, which are further augmented using Eqs.~(\ref{eq:cov})--~(\ref{eq:augmentation}). Source- and target-domain prototypes are then constructed, and the model is updated by minimizing the bi-directional adaptation loss in Eq.~(\ref{eq:overall_loss}). After adaptation, the target prototypes are recalculated using the adapted representation extractor, and the query samples are classified according to their Euclidean distances to these prototypes. The overall procedure is summarized in Algorithm~\ref{alg:implementation}.
\begin{algorithm}
\caption{Implementation procedure of the proposed DT-assisted few-shot fault diagnosis method}
\label{alg:implementation}
\begin{algorithmic}[1]
\REQUIRE DT source dataset $\mathcal{D}_s$; target support set $\mathcal{S}$;
target query set $\mathcal{Q}$; feature extractor $f$ with parameter $\theta$
\ENSURE Predicted labels of $\mathcal{Q}$

\STATE \textbf{Offline meta-training:}
\FOR{each source-domain meta-training episode}
    \STATE Sample source support and query sets from $\mathcal{D}_s$
    \STATE Extract features using the multi-periodicity learning network
    \STATE Construct source class prototypes using Eq.~(\ref{eq:proto})
    \STATE Compute $\mathcal{L}_s$ using Eq.~(\ref{eq:protoloss}) and update $\theta$
\ENDFOR

\STATE \textbf{Test-time adaptation:}
\STATE Extract target support features using the meta-trained $f$
\STATE Generate covariance-guided augmented features using Eqs.~(\ref{eq:cov})--(\ref{eq:augmentation})
\FOR{each adaptation epoch}
    \STATE Construct source-domain prototypes $\mathbf{p}^{s}_{n}$
    \STATE Construct target-domain prototypes $\mathbf{p}^{t}_{n}$
    \STATE Compute target-to-source loss $\mathcal{L}_{ts}$ using Eq.~(\ref{eq:tsa})
    \STATE Compute source-to-target loss $\mathcal{L}_{st}$ using Eq.~(\ref{eq:sta})
    \STATE Update $\theta$ by minimizing
    $\mathcal{L}_{t}$ of Eq.~(\ref{eq:overall_loss})
\ENDFOR

\STATE \textbf{Target-domain inference:}
\STATE Recalculate target prototypes using the adapted $f$
\FOR{each query sample $\mathbf{x}_{q}\in\mathcal{Q}$}
    \STATE Extract $f(\mathbf{x}_{q})$
    \STATE Assign $\mathbf{x}_{q}$ to the nearest target-domain prototype
\ENDFOR

\RETURN Predicted query labels
\end{algorithmic}
\end{algorithm}
}

\section{Experimental Verification}
\label{sec:Experiment}
\subsection{Experiment Setup}
In the experiment, current signals of the physical motor are collected with a sampling frequency of 5120 Hz. Since the sampling frequency of the simulated signals is twice that of the measured signals, the simulated data were downsampled by a factor of two to ensure consistency. For sample construction, segments with a duration of 0.2 s were extracted to form individual samples, resulting in the sample length of 1024.

To evaluate the few-shot fault diagnosis capability of the proposed method, four few-shot scenarios, namely 1-shot, 3-shot, 5-shot, and 10-shot, were designed for each working condition. {In each task, 15 query samples per class were used to evaluate the model performance. It should be emphasized that the query samples were strictly held out during test-time adaptation. Only the target-domain support samples and source-domain samples were used to update the model parameters, while the query set was used solely for evaluating the adapted model.} To obtain reliable results, 20 randomly generated tasks were constructed for each scenario, and the average accuracy was reported. Furthermore, all experiments were repeated five times with different random seeds to mitigate the influence of randomness.

For the model configuration, the kernel channel number in the MSCNN module was set to 32, and the top-5 dominant periods were selected for multi-periodicity modeling. Three ResBlocks with channel sizes of 64, 128, and 256 were employed, and each block contained two residual blocks. {The optimization of the proposed framework is conducted in two stages. During meta-training, the network parameters are updated by minimizing Eq.~(\ref{eq:protoloss}). The model is trained for 200 iterations with a learning rate of $1\times10^{-3}$. For each target task, the meta-trained parameters are subsequently used as the initialization for test-time adaptation. A batch of 40 source-domain samples was used for adaptation, and the feature augmentation factor was set to 3, meaning that three additional augmented samples were generated for each support feature. The representation extractor is then updated for 10 epochs by minimizing Eq.~(\ref{eq:overall_loss}) with a learning rate of $1\times10^{-4}$. The Adam optimizer was used in both meta-training and test-time adaptation. Among these hyperparameters, the number of dominant periods $k$ directly controls the amount of periodic information modeled by the network and is therefore treated as a major method-specific hyperparameter, whose influence will be systematically evaluated in Section~\ref{subsec:dominant_k}. All hyperparameters and experimental settings were fixed for all few-shot settings and working conditions. The key hyperparameters are listed in Table~\ref{tab:hyperparameters}.
}
\begin{table}
	
	\centering
	\caption{Key hyperparameters of the proposed method.}
	\label{tab:hyperparameters}
	\setlength{\tabcolsep}{5pt}
	\renewcommand{\arraystretch}{1.4}
	\begin{tabular}{lll}
		\hline
		Stage & Hyperparameter & Value \\
		\hline
		\multirow{3}{*}{Feature extraction}
		& Number of dominant periods $k$ & 5 \\
		& MSCNN channels & 32 \\
		& ResBlock channels & 64/128/256 \\
		\hline
		\multirow{2}{*}{Meta-training}
		& Learning rate & $1\times10^{-3}$ \\
		& Training iterations & 200 \\
		\hline
		\multirow{4}{*}{Test-time adaptation}
		& Learning rate & $1\times10^{-4}$ \\
		& Source batch size & 40 \\
		& Adaptation epochs & 10 \\
		& Augmentation factor $K_A$ & 3 \\
		\hline
	\end{tabular}
\end{table}

\subsection{Comparative Study}
\begin{figure*}
	\centering
	\subfloat[]{
		\includegraphics[width=0.3\linewidth]{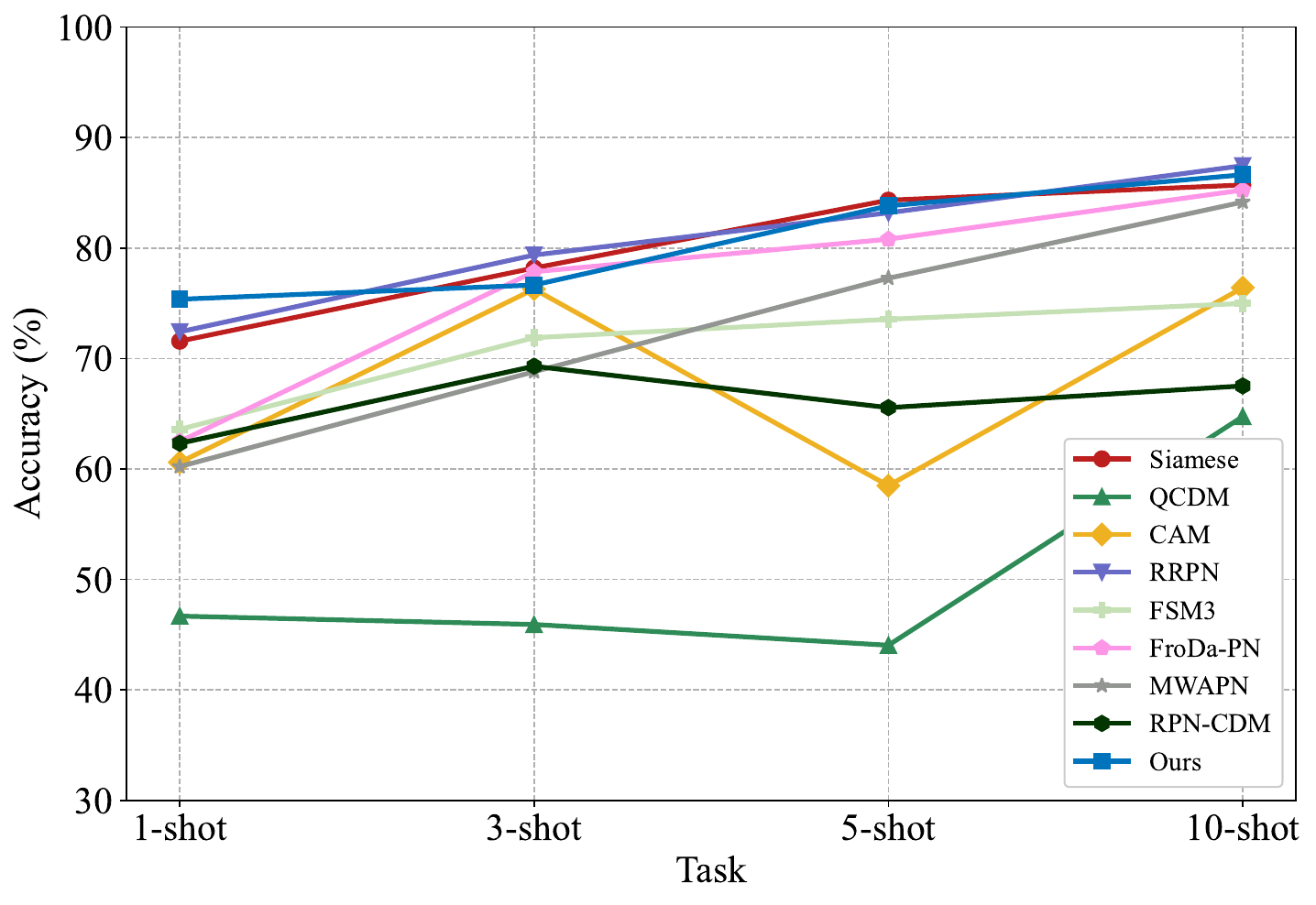}
	}
	\subfloat[]{
		\includegraphics[width=0.3\linewidth]{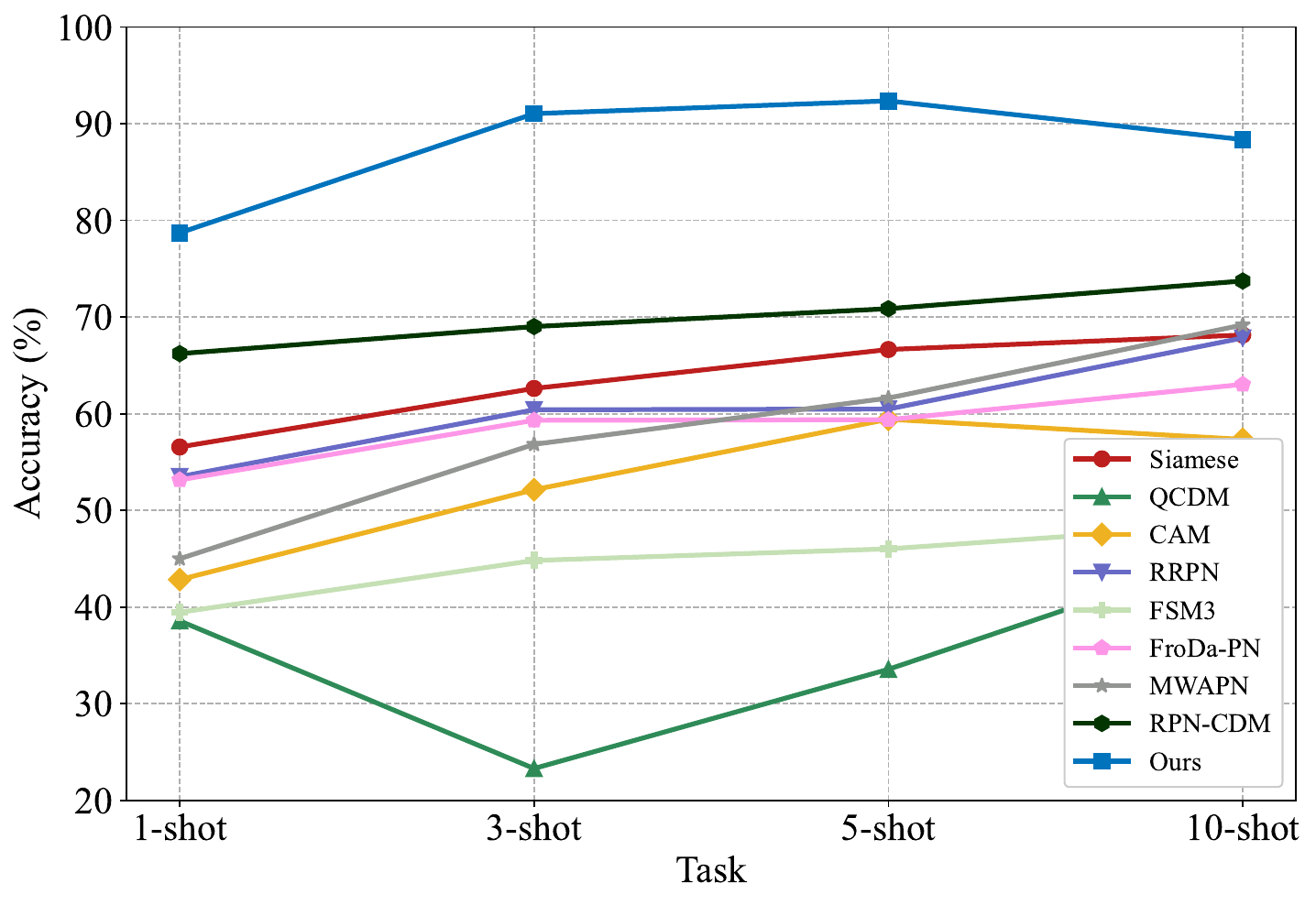}
	}
	\subfloat[]{
		\includegraphics[width=0.3\linewidth]{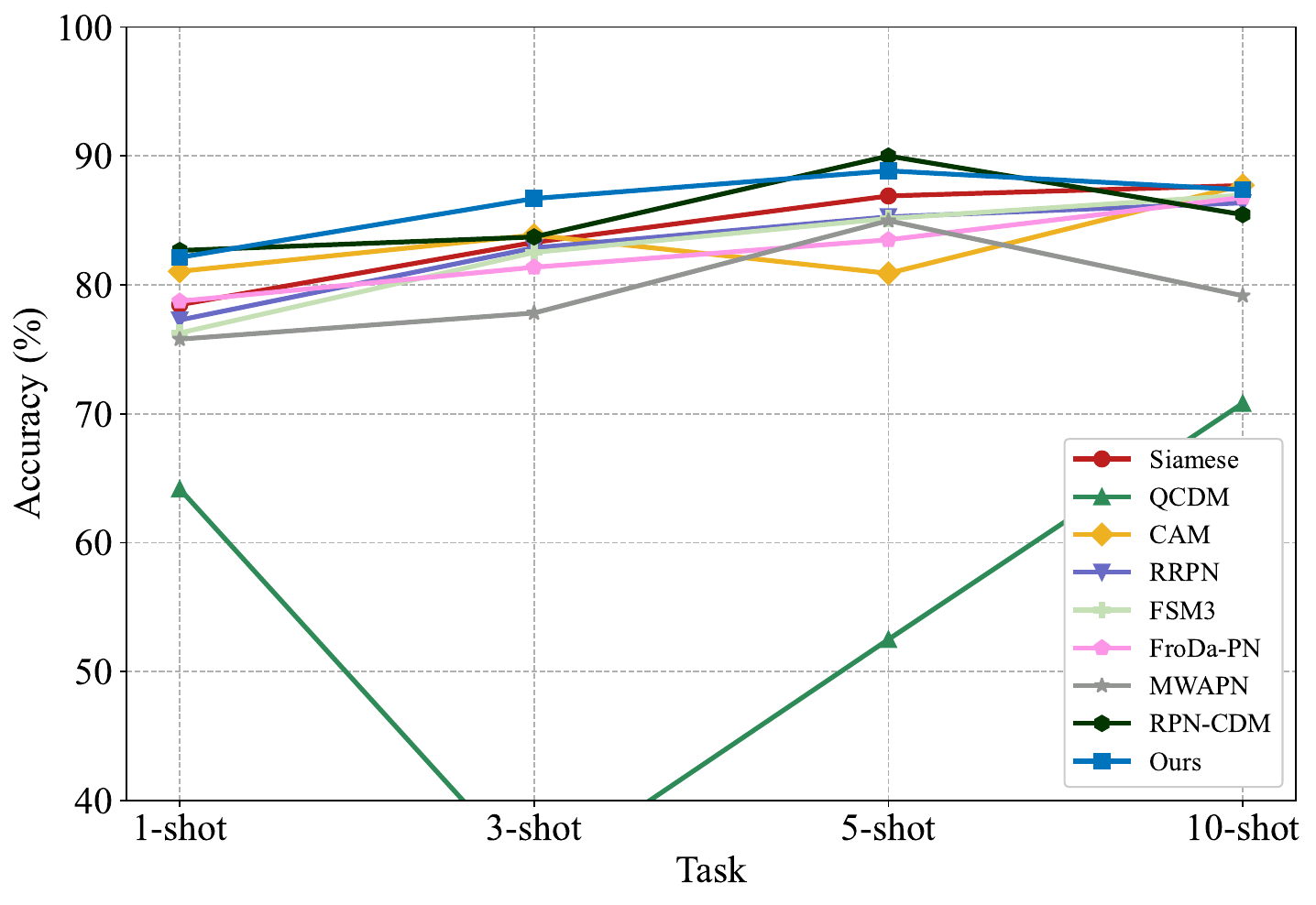}
	}
	\caption{Diagnosis accuracy of different methods in various few-shot settings. (a) 1200 r/min. (b) 2400 r/min. (c) 2700 r/min.}
	\label{fig:comparison}
\end{figure*}
{Eight representative few-shot learning methods were introduced for comparison to comprehensively evaluate the proposed method, including both classical approaches and recent state-of-the-art few-shot fault diagnosis methods:}
\begin{enumerate}[leftmargin=*]
	\item\textit{Siamese}~\cite{Siamese}: The Siamese Network is a classical metric-learning-based framework that learns a similarity function between pairs of samples through twin networks.
	\item\textit{QCDM}~\cite{QCDM}: It introduces a query-centric distance modulation (QCDM) strategy, which dynamically adjusts the distances between query samples and class prototypes.
	\item\textit{CAM}~\cite{CAM}: It focuses on reducing the distribution discrepancy between different classes by introducing a category alignment mechanism (CAM).
	\item\textit{RRPN}~\cite{RRPN}: The reweighted regularized ProtoNet (RRPN) introduces prototype reweighting and regularization strategies to assign adaptive weights to prototypes.
	\item\textit{FSM3}~\cite{FSM3}: It is a feature-space metric-based meta-learning (FSM3) framework tailored for fault diagnosis by constructing a more discriminative feature embedding space.
	\item \textit{FroDa-PN}~\cite{FroDa-PN}: A front-door adjustment-based prototype network (FroDa-PN) designed for cross-component few-shot fault diagnosis, which introduces causal adjustment to mitigate the influence of unobservable confounders.
	\item \textit{MWAPN}~\cite{MWAPN}: A multi-scale wavelet attention prototype network (MWAPN) developed for cross-domain few-shot fault diagnosis, which enhances multi-scale feature representation and prototype construction under distribution shifts.
	\item \textit{RPN-CDM}~\cite{RPN-CDM}: A retrospective prototype network with a center difference measure (RPN-CDM) for cross-machine few-shot fault diagnosis, which improves prototype representation and similarity measurement by exploiting class-center differences.
\end{enumerate}
All the methods were trained using the DT simulated data and verified with the few-shot measurement samples.

{
As illustrated in Fig.~\ref{fig:comparison}, the proposed method generally maintains high diagnostic accuracy across different working conditions and few-shot settings. Under 1200~r/min, the proposed method achieves 75.37\% accuracy in the challenging 1-shot setting and remains competitive as more support samples become available. Several recent methods, including FroDa-PN and MWAPN, also exhibit improved performance with increasing shots, indicating their effectiveness in conventional few-shot diagnosis. Under 2400~r/min, the superiority of the proposed method becomes particularly pronounced. This result indicates that few-shot prototype learning alone may be insufficient when a considerable simulation-to-measurement discrepancy exists, while the proposed test-time adaptation can effectively exploit the DT-derived diagnostic knowledge to adapt the representation to the physical asset. For the 2700~r/min condition, most recent few-shot methods achieve relatively strong performance. In particular, RPN-CDM slightly exceeds the proposed method in the 1-shot and 5-shot tasks. Nevertheless, the proposed method maintains consistently high accuracy across all four tasks. This suggests that the proposed method provides robust performance across different support-set sizes.

Although the state-of-the-art approaches improve few-shot diagnosis through advanced prototype construction or cross-domain representation learning, they primarily address knowledge transfer across physical machines, components, or operating conditions, and they rely heavily on reliable target support samples to provide fault information of the target asset. In the proposed method, the stable diagnostic knowledge learned from abundant DT data and the asset-specific information provided by the few target samples are jointly exploited by model adaptation. Consequently, the model maintains more consistent performance when the available target samples are extremely limited.
}

\subsection{Ablation Study}
\begin{table*}
	\centering
	\caption{Accuracy of ablation study in different scenarios (\%). $K$s denotes $K$-shot.}\label{tab:ablation}
	\setlength{\tabcolsep}{5pt}
	\renewcommand{\arraystretch}{1.4}
	\begin{tabular}{c|ccccc|ccccc|ccccc}
		\hline
		\multirow{2}{*}{Method} & \multicolumn{5}{c|}{1200 r/min}               & \multicolumn{5}{c|}{2400 r/min}               & \multicolumn{5}{c}{2700 r/min}               \\
		\cline{2-16}
                        & 1s & 3s & 5s & 10s & Average & 1s & 3s & 5s & 10s & Average & 1s & 3s & 5s & 10s & Average \\\hline
\textit{Baseline}                & 55.42  & 75.83  & 78.43  & 82.83   & 73.13   & 31.18  & 37.00  & 39.87  & 42.02   & 37.52   & 77.38  & 82.35  & 87.90  & 86.85   & 83.62   \\
\textit{w/o MPL}                 & 68.27  & 77.08  & 74.45  & 79.87   & 74.92   & 51.93  & 54.02  & 58.77  & 64.28   & 57.25   & 79.02  & 83.17  & 83.42  & 86.72   & 83.08   \\
\textit{MSCNN}                   & 73.47  & \textbf{81.15}  & 78.77  & 79.90   & 78.32   & 54.75  & 56.17  & 60.48  & 65.65   & 59.26   & 78.03  & 85.05  & 86.48  & 86.82   & 84.10   \\
\textit{w/o TTA}                 & 73.90  & \underline{78.83}  & 80.98  & \textbf{86.72}   & 80.11   & \textbf{79.45}  & 89.85  & 89.73  & 85.80   & 86.21   & \underline{82.55}  & \textbf{86.90}  & 87.08  & 87.23   & 85.94   \\
 \textit{w/o S-TA} &  74.50 &  75.62 &  83.18 &  86.00 &  79.83 &  79.15 &  \underline{90.53} &  91.75 &  \underline{88.60} &  87.51 &  82.43 &  86.61 &  88.28 &  87.20 &  86.13 \\
 \textit{w/o T-SA} &  \underline{74.77} &  76.63 &  \underline{83.63} &  86.42 &  \underline{80.36} &  78.58 &  90.50 &  92.22 &  88.28 &  87.40 &  82.08 &  86.62 &  \underline{88.98} &  \underline{87.37} &  \underline{86.26} \\
\textit{w/o CGA}                 & 73.28  & 73.47  & 83.55  & 85.38   & 78.92   & \underline{79.27}  & 90.42  & \textbf{92.77}  & \textbf{89.20}   & \textbf{87.91}   & \textbf{82.62}  & 85.37  & \textbf{89.07}  & 87.28   & 86.08   \\
\textit{Proposed}                & \textbf{75.37}  & 76.65  & \textbf{83.82}  & \underline{86.63}   & \textbf{80.62}   & 78.72  & \textbf{91.05}  & \underline{92.37}  & 88.37   & \underline{87.62}   & 82.13  & \underline{86.70}  & 88.85  & \textbf{87.38}   & \textbf{86.27}\\  
	\hline
	\end{tabular}
\end{table*}
To validate the effectiveness of each component in the proposed method, several variants were designed by removing individual components for ablation studies. The \emph{w/o TTA} represents the method without the test-time adaptation stage. {To further investigate the necessity of bi-directional prototype anchoring, two unidirectional variants were introduced. Specifically, \emph{w/o S-TA} removes the source-to-target adaptation loss $\mathcal{L}_{st}$ and retains only the target-to-source adaptation, whereas \emph{w/o T-SA} removes the target-to-source adaptation loss $\mathcal{L}_{ts}$ and retains only the source-to-target adaptation.} In \emph{w/o CGA}, the covariance-guided augmentation process is removed while the adaptation process is retained. The \textit{w/o MPL} represents a method, in which the multi-periodicity learning module is removed and a simple 1D convolutional layer is utilized, to verify the effectiveness of multi-periodicity learning. Meanwhile, the \textit{MSCNN} is designed by replacing the multi-periodicity module with 1D MSCNN instead of the 2D MSCNN in our network. The \textit{Baseline} method further removes the test-time adaptation process from \textit{w/o MPL}.

The ablation results under various working conditions and few-shot settings are summarized in Table~\ref{tab:ablation}. Overall, the proposed method achieves the highest or highly competitive accuracy across most scenarios and obtains the best average results in two conditions, indicating that each designed component contributes to the overall performance improvement.

First, the effectiveness of the multi-periodicity learning module can be verified by comparing the proposed method and the \textit{w/o MPL} method. The \textit{w/o MPL} method achieves relatively limited diagnostic accuracy, particularly under the 2400 r/min condition where around 30\% accuracy degradation is observed. Accuracy drops of \textit{w/o MPL} occur in almost all the scenarios, demonstrating that explicitly modeling the multi-periodic characteristics of current signals can enhance the discriminative ability of extracted features. Although \textit{MSCNN} achieves noticeable improvements compared with the \textit{Baseline} and \textit{w/o MPL}, its performance is still inferior to the proposed method in most cases. It suggests that simple multiscale convolution is insufficient to capture the intrinsic characteristics of current signals, whereas the proposed multi-periodicity learning can result in significant performance improvements.

Compared with \textit{w/o TTA}, introducing the adaptation mechanism in the proposed method makes the diagnostic performance more stable across various few-shot tasks, and average accuracy improvements are achieved under all the three working conditions. This indicates that the adaptation process helps mitigate the distribution discrepancy between virtual and physical domains, thereby improving the few-shot performance on the target asset.

{ Removing either the source-to-target adaptation or target-to-source adaptation results in drop of average accuracies under all three operating conditions. These results reveal the complementary roles of the two adaptation directions. Target-to-source adaptation exploits the relatively stable DT-derived prototypes to regularize the scarce target samples, whereas source-to-target adaptation introduces information from the physical-domain prototypes to compensate for the simulation-to-measurement discrepancy. Although either unidirectional strategy can already provide effective adaptation, combining both directions consistently provides the best average performance, indicating that mutual prototype anchoring yields a more balanced and robust alignment between the two domains.}

Furthermore, the comparison between \textit{w/o CGA} and the proposed method verifies the contribution of the covariance-guided augmentation strategy. Removing the augmentation process leads to a slight decrease in diagnostic accuracy in most tasks. This result suggests that the strategy enables the model to construct more reliable class prototypes under limited labeled samples and results in more robust performance.

\subsection{Discussion}
\subsubsection{Confusion matrices}
\begin{figure}
	\centering
	\subfloat[]{
		\includegraphics[width=0.22\linewidth]{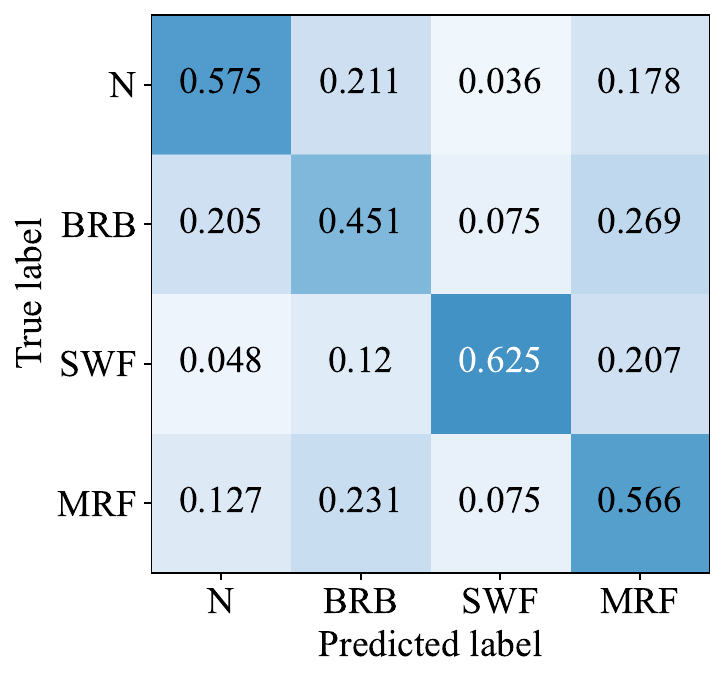}
	}
	\subfloat[]{
		\includegraphics[width=0.22\linewidth]{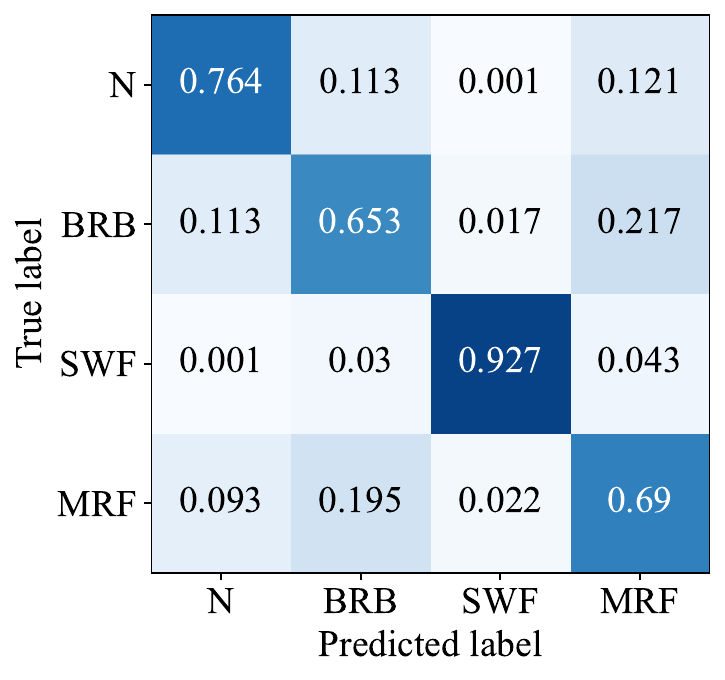}
	}
	\subfloat[]{
		\includegraphics[width=0.22\linewidth]{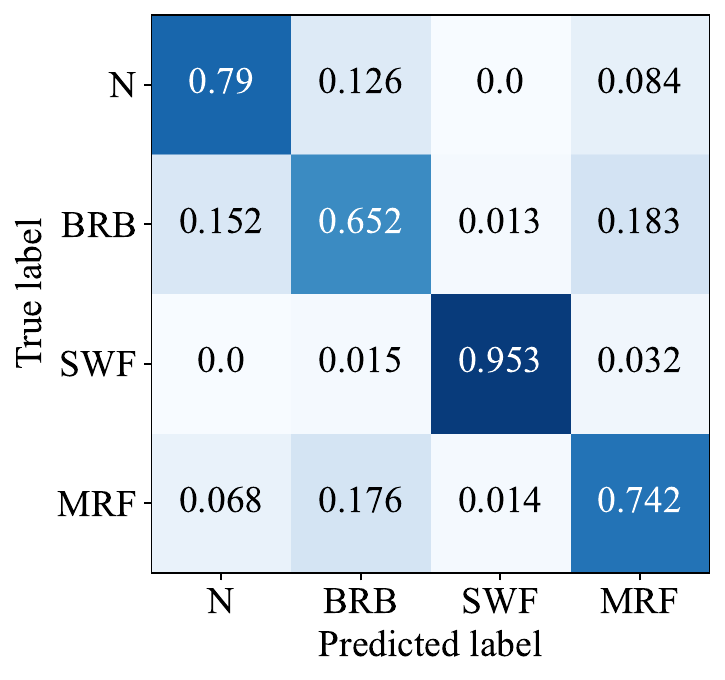}
	}
	\subfloat[]{
		\includegraphics[width=0.22\linewidth]{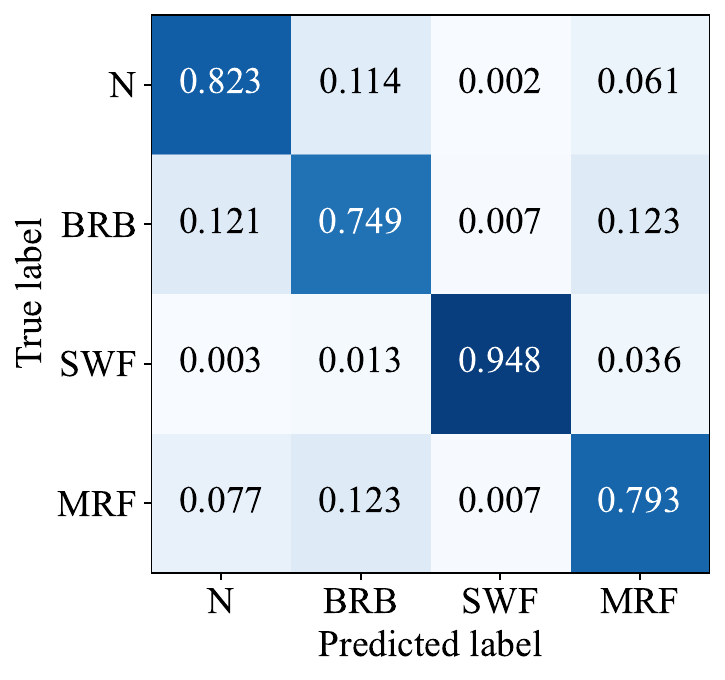}
	}
	\hfill
	\subfloat[]{
		\includegraphics[width=0.22\linewidth]{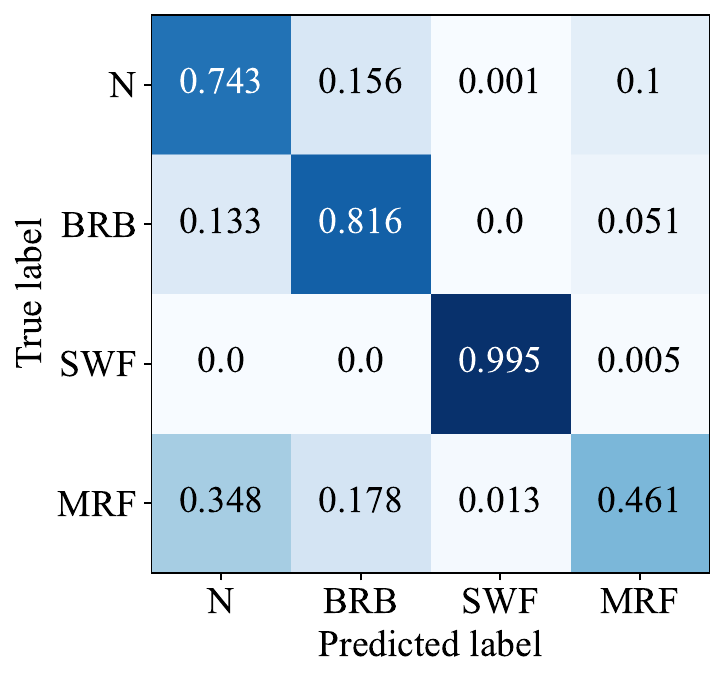}
	}
	\subfloat[]{
		\includegraphics[width=0.22\linewidth]{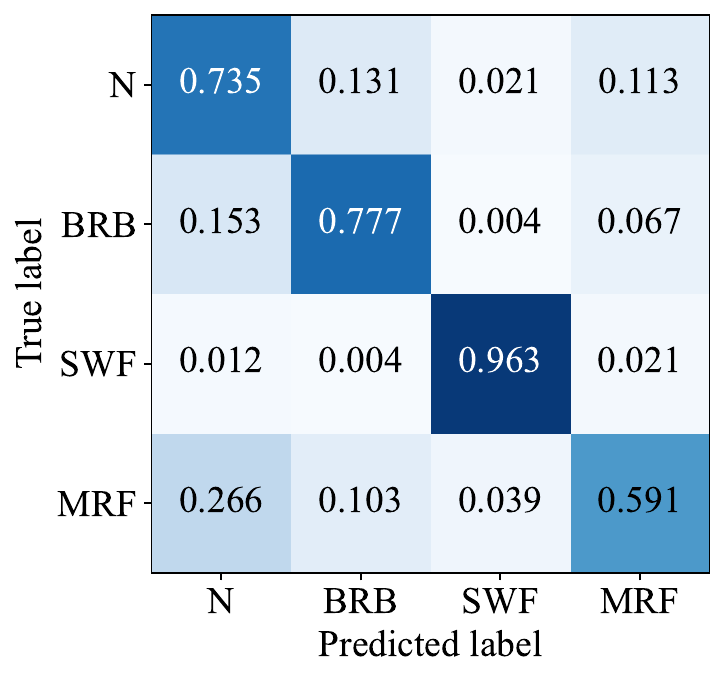}
	}
	\subfloat[]{
		\includegraphics[width=0.22\linewidth]{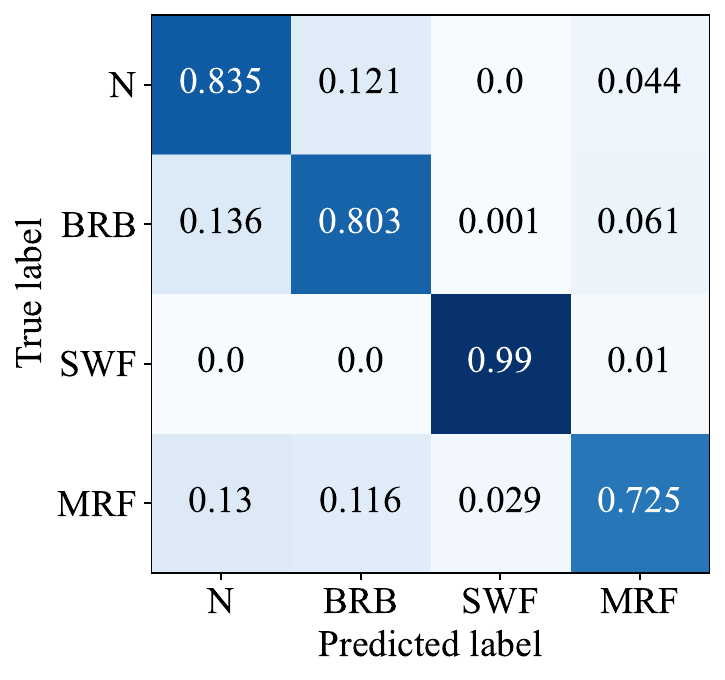}
	}
	\subfloat[]{
		\includegraphics[width=0.22\linewidth]{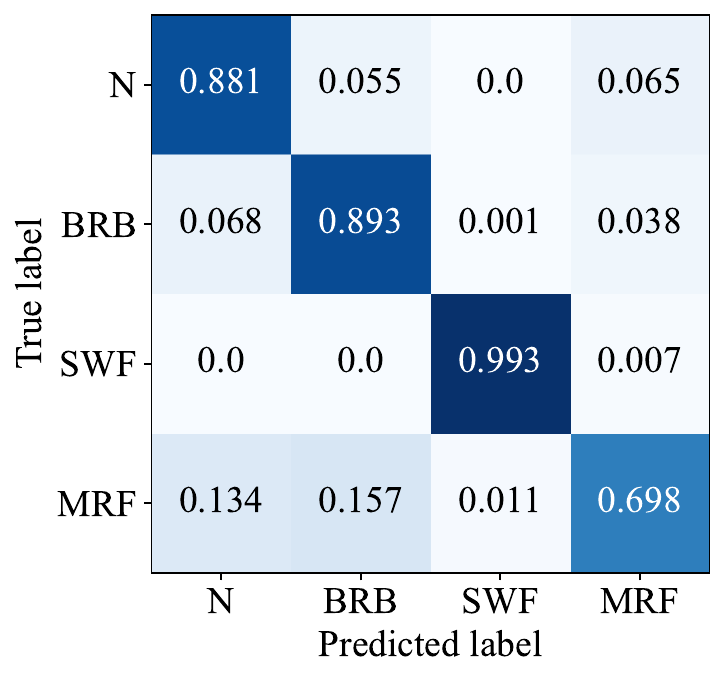}
	}
	\caption{Confusion matrices of the \textit{Baseline} method (a)-(d) and the proposed method (e)-(h) in scenarios from 1-shot to 10-shot under the condition of 1200 r/min}
	\label{fig:matrix}
\end{figure}

The confusion matrices of the \textit{Baseline} and the proposed method under the first working condition are depicted in Fig.~\ref{fig:matrix}. The proposed method consistently outperforms the \textit{Baseline} method for all the categories in almost all the few-shot settings. Both methods can recognize the SWF state well due to the obvious imbalance in three-phase signals in time domain. Although the proposed method achieves better performance, it struggles to recognize the MRF state, which is difficult to identify from the time-domain signals as well.

{
\subsubsection{Feature visualization}
To provide an intuitive understanding of the adaptation effect, feature visualizations for the 5-shot tasks under the three working conditions are shown in Fig.~\ref{fig:tsne_all_conditions}. In each case, the source-domain features/prototypes and the target-domain features before and after adaptation are visualized using t-SNE. After adaptation, the target-domain features become more consistent with the source-domain class structure while maintaining clear class separability in the target domain.
\begin{figure}
	\centering
	\subfloat[]{
		\includegraphics[width=0.8\linewidth]{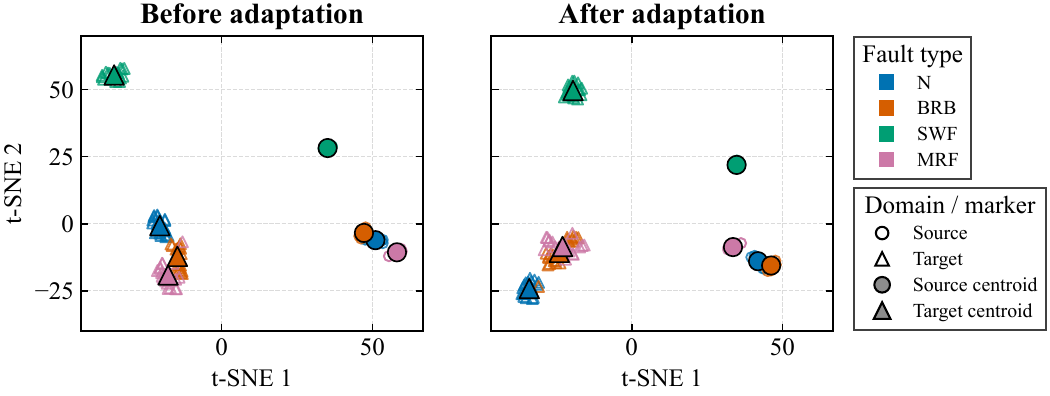}
	}\hfill
	\subfloat[]{
		\includegraphics[width=0.8\linewidth]{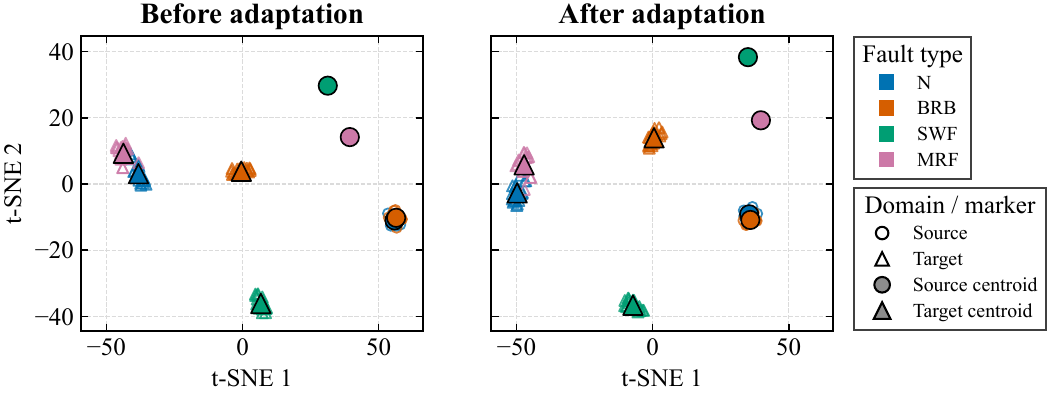}
	}
	\hfill
	\subfloat[]{
		\includegraphics[width=0.8\linewidth]{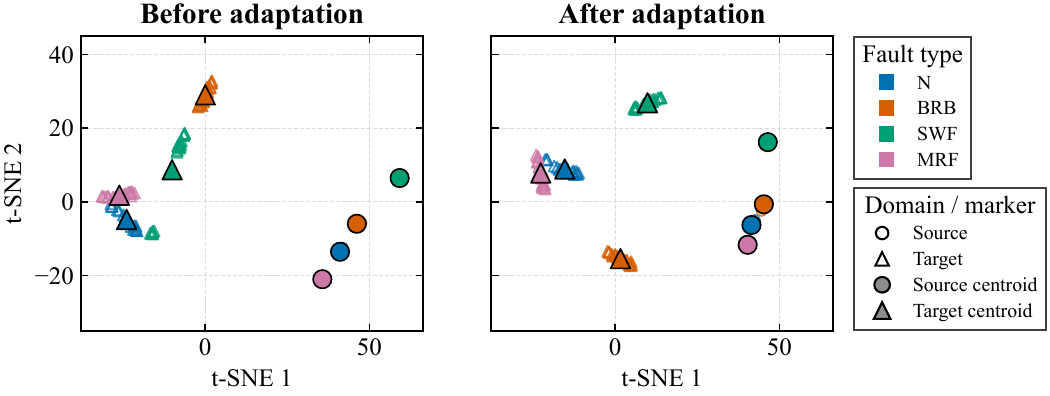}
	}
	\caption{t-SNE visualizations of the source-domain features/prototypes and the target-domain features before and after adaptation under the 5-shot tasks of three working conditions: (a) 1200~r/min, (b) 2400~r/min, and (c) 2700~r/min. Open circles and open triangles denote individual source-domain and target-domain samples, respectively, while filled circles and filled triangles denote the corresponding class centroids.}
	\label{fig:tsne_all_conditions}
\end{figure}

It should be noted that these t-SNE plots are mainly used for qualitative illustration. Since t-SNE is a nonlinear dimensionality-reduction method, the 2D embeddings do not necessarily preserve the exact distances in the original high-dimensional feature space. Therefore, to quantitatively evaluate the alignment effect, the Euclidean distances between the source-domain and target-domain class centroids were further computed in the original feature space. As shown in Table~\ref{tab:centroid_distance_summary}, mean source--target centroid distances across four categories reduce after test-time adaptation under all the three working conditions. Overall, the average centroid distance decreases by 31.45\% after adaptation. These results confirm that the proposed adaptation effectively reduces the simulation-to-measurement discrepancy under different working conditions.}
\begin{table}
	\centering
	\setlength{\tabcolsep}{12pt}
	\renewcommand{\arraystretch}{1.4}
	\caption{Mean source--target centroid distances in the original feature space before and after adaptation under different working conditions (5-shot).}
	\label{tab:centroid_distance_summary}
	
	\begin{tabular}{cccc}
		\hline
		Condition & Before & After & Reduction \\
		\hline
		1200 r/min & 11.21 & 10.05 & 10.45\% \\
		2400 r/min & 29.06 & 20.92 & 28.12\% \\
		2700 r/min & 45.58 & 27.88 & 38.83\% \\
		\hline
		Overall & 28.62 & 19.62 & 31.45\% \\
		\hline
	\end{tabular}
\end{table}

{
It is also worth noting that the purpose of the proposed adaptation is not to force the source and target domains to become completely overlapped within a few test-time update steps. Instead, the method aims to reduce the class-wise domain discrepancy while preserving a discriminative structure in the target domain. Moreover, the final few-shot prediction is performed based on the target-domain prototypes. Therefore, complete overlap between the two domains is not a necessary condition for accurate diagnosis.
}

\subsubsection{Sensitivity of number of dominant periods}\label{subsec:dominant_k}
\begin{figure}
	\centering
	\includegraphics[width=0.6\linewidth]{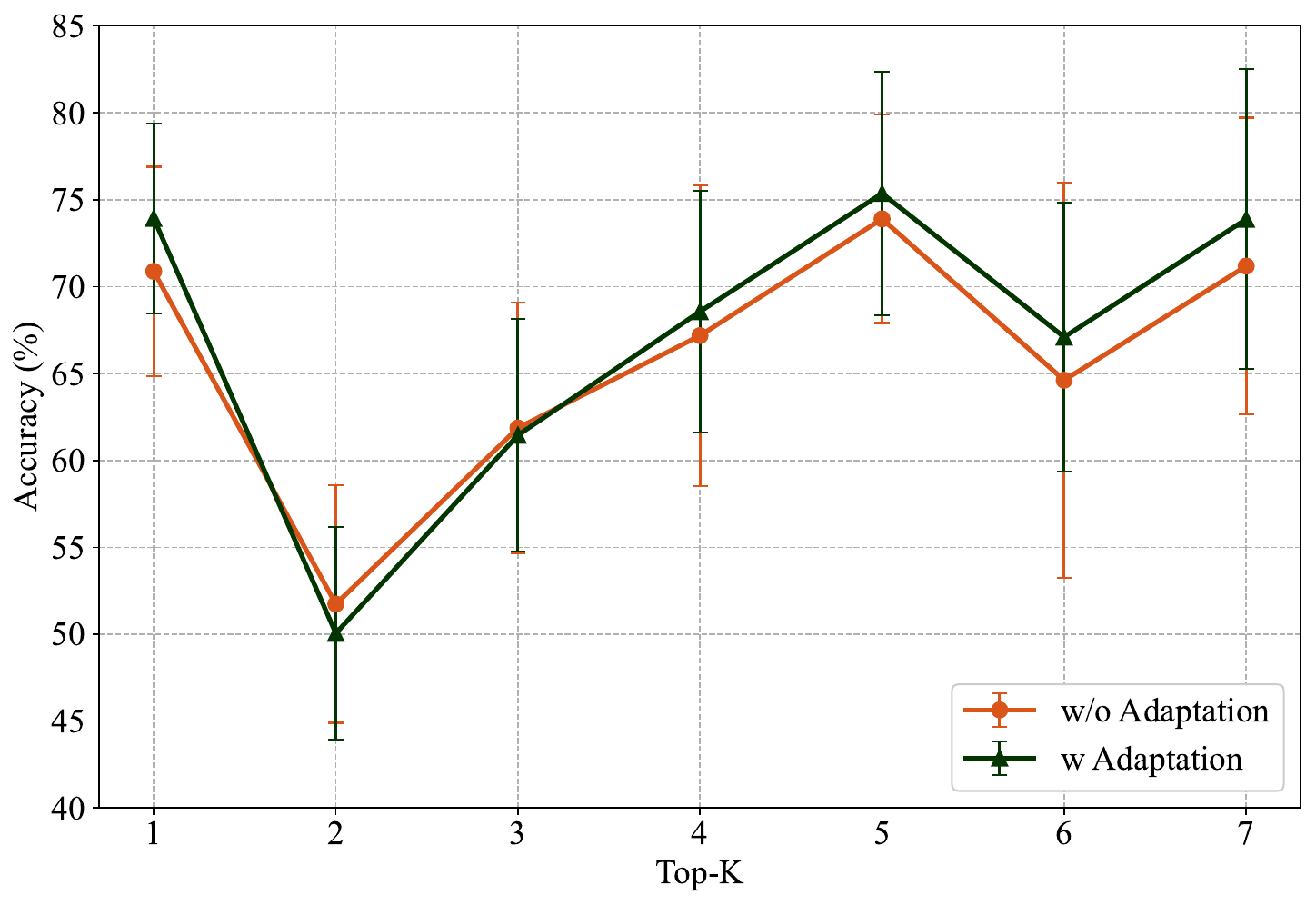}
	\caption{Model accuracy with different top-$k$ values under the condition of 1200 r/min.}
	\label{fig:topk}
\end{figure}

As the number of dominant periods in the multi-periodicity learning module is a critical hyperparameter, experiments were conducted to investigate its influence on the model performance. Specifically, the model performance with and without test-time adaptation was evaluated under different top-$k$ settings ranging from 1 to 7 under the first working condition. As illustrated in Fig.~\ref{fig:topk}, both methods exhibit a similar performance trend as the value of $k$ varies. When the number of dominant periods increases from 1 to 2, the diagnostic accuracy drops significantly. This phenomenon can be attributed to the fact that introducing a small number of additional frequency components may incorporate non-dominant or noisy periodic components, which weakens the discriminative characteristics of the reconstructed signals. As the value further increases from 2 to 5, the diagnostic accuracy gradually improves. The best performance for both methods is achieved when $k=5$, suggesting that five dominant periods provide a suitable balance between capturing sufficient periodic information and avoiding excessive redundant components. When $k$ continues to increase beyond this value, the performance no longer improves and even shows a slight decline. Therefore, an appropriate selection of dominant periods is essential for effective intrinsic characteristic modeling.

{
\subsubsection{Sensitivity of query size}
To investigate whether the reported performance is sensitive to the query-set size, additional experiments were conducted by varying the number of query samples per class to 5, 10, and 20, in addition to the default setting of 15. All other settings, including the support set and the test-time adaptation procedure, were kept unchanged.

As shown in Table~\ref{tab:query_size}, the proposed method maintains highly consistent performance across different query-set sizes. The overall average accuracies with 5, 10, 15, and 20 query samples per class are 85.00\%, 84.96\%, 84.84\%, and 84.82\%, respectively, with a maximum difference of only 0.18 percentage points. Similarly, the average accuracy varies only slightly within each working condition. No monotonic performance change is observed with increasing query-set size. This behavior is consistent with the design of the proposed framework, since the query samples are not involved in the test-time adaptation. Therefore, varying its size mainly affects the statistical estimation of diagnostic accuracy rather than the adapted model itself. The consistent results across different query-set sizes further demonstrate the robustness of the proposed method and confirm that the reported performance is not dependent on a particular query-set configuration.
\begin{table*}
	\centering
	
	\caption{Diagnostic accuracy (\%) with different numbers of query samples per class.}\label{tab:query_size}
	\setlength{\tabcolsep}{5pt}
	\renewcommand{\arraystretch}{1.4}
	\begin{tabular}{c|ccccc|ccccc|ccccc}
		\hline
		\multirow{2}{*}{Query size}
		& \multicolumn{5}{c|}{1200 r/min}
		& \multicolumn{5}{c|}{2400 r/min}
		& \multicolumn{5}{c}{2700 r/min} \\
		\cline{2-16}
		& 1s & 3s & 5s & 10s & Average
		& 1s & 3s & 5s & 10s & Average
		& 1s & 3s & 5s & 10s & Average\\
		\hline
		5
		& 74.15 & 78.50 & 82.65 & 87.40 & 80.67
		& 79.05 & 92.35 & 92.30 & 87.40 & 87.77
		& 81.85 & 86.25 & 89.45 & 88.70 & 86.56\\

		10
		& 74.60 & 77.78 & 82.95 & 86.65 & 80.49
		& 79.98 & 90.38 & 92.15 & 88.35 & 87.71
		& 82.73 & 86.28 & 89.15 & 88.53 & 86.67\\

		15
		& 75.37 & 76.65 & 83.82 & 86.63 & 80.62
		& 78.72 & 91.05 & 92.37 & 88.37 & 87.62
		& 82.13 & 86.70 & 88.85 & 87.38 & 86.27\\

		20
		& 74.21 & 77.71 & 83.84 & 87.15 & 80.73
		& 79.09 & 91.08 & 92.45 & 88.26 & 87.72
		& 81.74 & 86.08 & 89.25 & 87.01 & 85.69\\
		\hline
	\end{tabular}
\end{table*}
}

{
\subsubsection{Practical applicability}
It should be emphasized that the current framework targets an asset-specific DT-to-physical adaptation scenario. It assumes that the DT provides physically meaningful fault signatures for the considered health conditions and that a small number of labeled physical samples are available for adaptation. The present formulation is therefore suitable for few-shot diagnosis of known fault categories on a physical asset associated with its DT, rather than direct zero-shot diagnosis of unknown faults or unrestricted cross-machine generalization.
}

\section{Conclusion}
\label{sec:Conclusion}
In this paper, a DT-assisted few-shot fault diagnosis framework is proposed to address the challenge of insufficient measurement data in practical industrial scenarios. Through DT modeling and simulated data generation, a framework involving meta-training in the virtual space and test-time adaptation in the physical space is proposed. In the adaptation stage, a twin-domain bi-directional prototype anchoring strategy is developed to update the model to alleviate data discrepancies between two domains. In addition, a covariance-guided augmentation mechanism is introduced to increase data diversity for target samples. Furthermore, a multi-periodicity feature learning module is designed to model the intrinsic periodic characteristics of current signals. Experimental results using a motor case study under multiple few-shot settings and working conditions demonstrate that the proposed method outperforms several representative few-shot fault diagnosis approaches. Ablation studies further validate the effectiveness of each proposed component. Future work will focus on exploring the model's robustness on more fault types and variable working conditions.

% References

\bibliographystyle{IEEEtran}
\bibliography{refs.bib}\ %IEEEabrv instead of IEEEfull

\end{document}